\documentclass{article}
\usepackage{arxiv}
\usepackage[utf8]{inputenc}
\usepackage[T1]{fontenc}
\usepackage[hyphens]{url}
\usepackage{graphicx}
\usepackage{natbib}
\setcitestyle{numbers,square,comma}
\usepackage{microtype}
\usepackage{booktabs}
\usepackage{array}
\usepackage{algorithm}
\usepackage{algorithmic}
\usepackage{amsmath}
\usepackage{pifont}
\usepackage{xcolor}
\usepackage{caption}
\usepackage{placeins}
\usepackage[skins,breakable]{tcolorbox}
\usepackage{fontawesome5}
\usepackage[hidelinks]{hyperref}

\newcommand{\code}[1]{\texttt{#1}}
\newcommand{\best}[1]{\textbf{#1}}

\newcommand{\cmark}{\textcolor{green!50!black}{\ding{51}}}
\newcommand{\xmark}{\textcolor{red!70!black}{\ding{55}}}
\newcommand{\pmark}{\textcolor{orange!85!black}{\(\triangle\)}}

\title{CRAFTS: Collaborative Role-Adaptive Fine-Tuning of LLM Agents for
Chemical Process Simulation}
\author{Ziyun Zhang$^{1,\dagger}$ \quad
Yuxin Lin$^{2,\dagger}$ \quad
Eldin Wee Chuan Lim$^{1,*}$ \quad
Xinghao Ding$^{2,*}$\\[0.65em]
{\normalfont\small $^1$Department of Chemical \& Biomolecular Engineering,
National University of Singapore}\\
{\normalfont\small $^2$Key Laboratory of Multimedia Trusted Perception and
Efficient Computing,}\\
{\normalfont\small Ministry of Education of China, Xiamen University, Xiamen,
Fujian, China}\\[0.35em]
{\normalfont\small $^\dagger$Equal contribution.
$^*$Corresponding authors.}\\[0.45em]
\href{https://github.com/Galigeigei-Z/CRAFTS-Multi-agent-for-Equation-oriented-PSE}
{\normalfont\faGithub\ \texttt{GitHub}}}
\date{}

\renewcommand{\headeright}{}
\renewcommand{\undertitle}{Preprint}
\renewcommand{\shorttitle}{\footnotesize CRAFTS: Collaborative Role-Adaptive
Fine-Tuning of LLM Agents for Chemical Process Simulation}
\hypersetup{
  pdftitle={CRAFTS: Collaborative Role-Adaptive Fine-Tuning of LLM Agents for Chemical Process Simulation},
  pdfauthor={Ziyun Zhang, Yuxin Lin, Eldin Wee Chuan Lim, Xinghao Ding},
  pdfkeywords={chemical process simulation, large language model agents, IDAES, Pyomo, LoRA}
}

\newcolumntype{P}[1]{>{\raggedright\arraybackslash}p{#1}}
\definecolor{traceblue}{HTML}{2B6CB0}
\definecolor{tracered}{HTML}{C53030}
\definecolor{tracegray}{HTML}{526171}
\definecolor{tracetitle}{HTML}{245F96}
\definecolor{tracepanel}{HTML}{F8FAFD}
\definecolor{tracebar}{HTML}{EAF2F9}
\definecolor{traceline}{HTML}{CFDCE8}
\definecolor{tracereceiver}{HTML}{5D6875}

\newtcolorbox{tracecasebox}[1]{%
  enhanced,
  breakable,
  width=0.97\textwidth,
  colback=tracepanel,
  boxrule=0pt,
  borderline={0.75pt}{0pt}{traceblue,dashed},
  arc=7pt,
  outer arc=7pt,
  boxsep=4pt,
  left=7pt,
  right=7pt,
  top=8pt,
  bottom=7pt,
  before skip=10pt,
  after skip=10pt,
  title={#1},
  coltitle=white,
  colbacktitle=tracetitle,
  fonttitle=\bfseries\scriptsize,
  attach boxed title to top left={xshift=13pt,yshift=-2mm},
  boxed title style={
    boxrule=0pt,
    arc=6pt,
    outer arc=6pt,
    left=7pt,
    right=7pt,
    top=1.5pt,
    bottom=1.5pt,
    colback=tracetitle
  }
}

\newtcolorbox{tracejsonbox}{%
  width=0.98\linewidth,
  colback=white,
  colframe=traceline,
  boxrule=0.45pt,
  arc=2pt,
  boxsep=3pt,
  left=5pt,
  right=5pt,
  top=4pt,
  bottom=4pt,
  before skip=5pt,
  after skip=3pt,
  fontupper=\ttfamily\tiny
}

\newcommand{\traceseparator}{%
  \par\noindent
  \textcolor{traceline}{\rule{\linewidth}{0.25pt}}%
  \par}

\newcommand{\runbar}[1]{%
  \noindent\colorbox{tracebar}{%
    \parbox{\dimexpr\linewidth-2\fboxsep\relax}{%
      \strut\textcolor{tracetitle}{%
        \textbf{ChE-knowledge-informed LangGraph trajectory}}\strut}}\par}

\newcommand{\tracestage}[2]{%
  \par\medskip
  \noindent\colorbox{tracebar}{%
    \parbox{\dimexpr\linewidth-2\fboxsep\relax}{%
      \strut\textcolor{tracetitle}{\textbf{Stage #1: #2}}\strut}}\par\smallskip}

\newcommand{\traceuser}[3]{%
  \traceseparator
  \noindent{\textcolor{tracered}{\textbf{#1}}}\enspace
  \textcolor{tracereceiver}{\emph{(to #2)}}: #3\par}

\newcommand{\traceagent}[3]{%
  \traceseparator
  \noindent{\textcolor{traceblue}{\textbf{#1}}}\enspace
  \textcolor{tracereceiver}{\emph{(to #2)}}:\par
  \noindent #3\par}

\newcommand{\tracetool}[3]{%
  \traceseparator
  \noindent{\textcolor{tracegray}{\textbf{#1}}}\enspace
  \textcolor{tracereceiver}{\emph{(to #2)}}:\par
  \noindent #3\par}

\newcommand{\casefigure}[3]{%
  \begin{figure}[p]
  \centering
  \includegraphics[width=0.96\textwidth]{#1}
  \caption{#2 Panels show (A) process evidence, (B) accepted TopologyIR,
  (C) the IDAES flowsheet view, and (D) terminal numerical evidence from the
  same trajectory.}
  \label{#3}
  \end{figure}
  \clearpage
}

\begin{document}

\maketitle

\begin{abstract}
Constructing an executable chemical-process model remains manually intensive.
Chemical engineers translate underspecified requests into coupled decisions
about unit operations, thermodynamics, streams,
specifications, degrees of freedom (DoF), initialization, solver repair, and
optimization; one error can invalidate the model. CRAFTS mirrors the staged
workflow of chemical engineers by decomposing simulation building into bounded
subtasks assigned to seven bounded roles, with
deterministic IDAES/Pyomo gates between stages. Given a natural-language request, process flowsheet diagram (PFD) evidence, and curated chemical-engineering knowledge, Input Understanding and Intent recover
requirements, constraints, and process semantics; visual, topology, and
specification specialists translate them into typed simulator contracts; and
Debug and Optimization support bounded repair and eligible optimization.
Fine-tuning is applied to the three schema-critical visual, topology, and
specification roles, while the remaining roles use untuned Qwen. The resulting
VisualGraphIR, TopologyIR, SpecIR, BuildPlan, and SolveReport expose unit, port,
thermodynamic, numerical, and execution decisions. Compatible constructors,
property packages, and runners are attached only after semantic artifacts pass
engineering gates. We introduce OpenIDAES-450, a 450-case IDAES process-
simulation dataset, and evaluate the complete seven-role LangChain/LangGraph
workflow through solve and eligible optimization on its frozen 82-case held-out
split. CRAFTS completes the prescribed validation and execution contract for for 91.5\% of cases and
achieves unit, stream, and directed-connection F1 scores of 0.815, 0.791, and
0.782. These results demonstrate the effectiveness of role specialization, typed intermediate representations, and deterministic engineering gates for reliable automated process-model construction.
\end{abstract}


\section{Introduction}

Large language models (LLMs) increasingly coordinate scientific tools
\cite{bran2024chemcrow,ruan2024llmrdf,li2026unim}, extract structured records from technical
text \cite{dagdelen2024structured}, and drive autonomous laboratory and discovery
workflows \cite{boiko2023autonomouschemical,ghareeb2026multiagentdiscovery}.
These systems show that LLMs can orchestrate scientific tools, but their targets
are experimental protocols, retrosynthesis, or extracted records rather than
executable process-simulation models.

Despite mature commercial simulators such as Aspen Plus/HYSYS,
chemical-process model construction still depends heavily on manual engineering
work and domain experience \cite{haydary2019chemical, ZHANG2026124158}. An engineer must interpret an
incomplete process need, organize unit operations and streams, choose compatible
thermodynamics and property packages, specify operating targets, close DoF, initialize recycles, and diagnose numerical failures. When a
model fails, the relevant cause may be distributed across flowsheet connections,
configuration choices, initialization procedures, and solver diagnostics,
making repair iterative and expert-intensive. These choices are tightly coupled:
a wrong endpoint, component alias, property boundary, or specification target
can invalidate downstream construction or solving. A plausible paragraph,
flowsheet sketch, or generated script is therefore not enough; an executable
model must satisfy structural, physical, numerical, and solver contracts
simultaneously.

Chemical engineers address this complexity through staged behavior. They recover requirements and constraints, assemble and inspect topology, close the numerical
model, materialize and initialize it, read diagnostics, and repair the failed
layer before optimization. While the CRAFTS decomposes model construction into distinct but dependent engineering subtasks,
assigns bounded responsibility to collaborating role-specialized LLM agents, and
passes typed artifacts through deterministic gates. This division of labor is
motivated by the different error surfaces of interpretation, topology,
specification, and execution, rather than by generic LLM orchestration. It
accelerates the engineer's iterative workflow without replacing engineering
judgment, while preserving the intermediate state and intervention points on
which that workflow relies.

\begin{figure*}[t]
\centering
\includegraphics[width=1\textwidth,height=0.50\textheight,keepaspectratio]{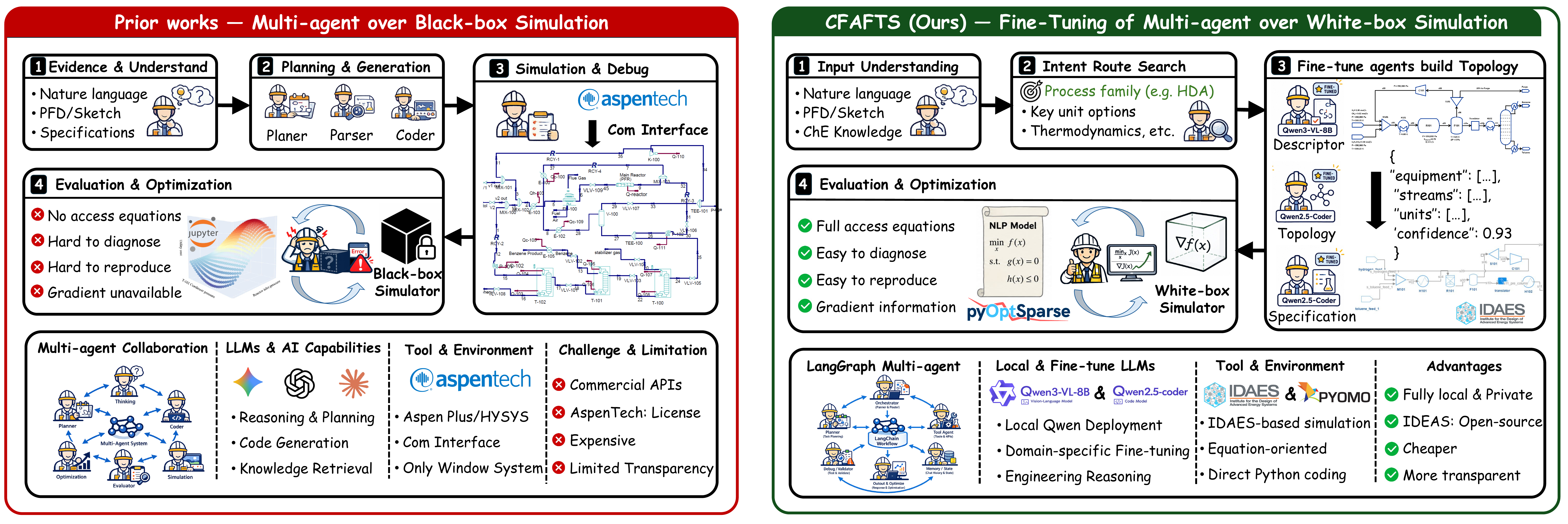}
\caption{From closed simulation workflows to inspectable equation-oriented
models. Left: conventional process simulation (e.g. Aspen Plus/HYSYS) relies on expert-driven
interaction with commercial simulators, where model construction and failure diagnosis often require
expert-driven interaction and uniform inspectable substrate for
agentic reasoning. Right: CRAFTS combines LoRA-adapted Qwen agents with an open
IDAES/Pyomo substrate. Agents generate typed engineering artifacts that are
compiled into executable models whose variables, constraints, equations, and
solver diagnostics remain inspectable for further validation, repair, and optimization.
The detailed multi-agent workflow and deterministic gates are presented in
Figure~\ref{fig:architecture}.}
\label{fig:representations}
\end{figure*}

This design follows process-systems engineering (PSE), where a flowsheet is an
equation-oriented model with unit-operation structure, thermophysical packages,
specifications, and solver state, not merely a diagram or paragraph
\cite{biegler1997systematic,grossmann2012advances}. Built on the open
IDAES/Pyomo substrate \cite{lee2021idaes,hart2011pyomo}, CRAFTS exposes equations, variables, constraints, and
solver state as inspectable and programmable objects, enabling localized
validation and repair. It materializes VisualGraphIR, TopologyIR, SpecIR,
BuildPlan, SolveReport, and optional optimization records before and during
execution. Figure~\ref{fig:representations} contrasts closed, expert-driven
simulation workflows with the open-substrate pipeline, while
Figure~\ref{fig:architecture} shows the typed handoffs, deterministic gates, and
bounded stage-specific repair.

\begin{table}[t]
\centering
\begin{tabular*}{\columnwidth}{@{\extracolsep{\fill}}P{0.25\columnwidth}cccc@{}}
\toprule
\textbf{Capability} & CePro.$^{1}$ & T2Sim.$^{2}$ & S2Sim.$^{3}$ & \textbf{Ours} \\
\midrule
Open equation substrate  & \xmark & \xmark & \xmark & \cmark \\
Structured artifacts & \pmark & \cmark & \cmark & \cmark \\
Role-specific fine-tuning & \xmark & \xmark & \xmark & \cmark \\
ChE knowledge  & \cmark & \pmark & \pmark & \cmark \\
Fail-closed checks & \pmark & \pmark & \cmark & \cmark \\
Process optimization & \cmark & \cmark & \xmark & \cmark \\
\bottomrule
\end{tabular*}
\caption{Demonstrated capabilities of contemporary process-simulation agent systems:
$^{1}$CeProAgents~\cite{yang2026ceproagents}, $^{2}$From Text to
Simulation~\cite{tian2026textsimulation}, and
$^{3}$Sketch2Simulation~\cite{bahamdan2026sketch2simulation}. \cmark{} first-class,
\pmark{} partial, \xmark{} functionality outside demonstrated scope.}
\label{tab:positioning}
\end{table}

CRAFTS makes four contributions:
\begin{itemize}

\item \textbf{Typed engineering formulation.}
CRAFTS converts requests and optional PFDs into explicit requirement, visual, topology, specification, build, and solve artifacts, exposing simulator-critical decisions before code generation.

\item \textbf{Role-specific adaptation.}
Seven bounded roles separate interpretation, graph recovery, numerical closure, diagnosis, and optimization; dedicated LoRA adapters are used only for the three schema-critical roles.

\item \textbf{Fail-closed executable workflow.}
Deterministic gates validate schema, topology, ports, thermodynamics, DoF, initialization, physical consistency, and solver status, while stage-local repair preserves previously validated state.

\item \textbf{Open benchmark and controlled evaluation.}
OpenIDAES-450 provides 450 auditable IDAES tasks, a frozen 82-case test split, and per-case artifacts and execution traces for reproducible comparison.

\end{itemize}

\section{Related Work}

\textbf{ChE representation and simulation.}
Natural language, PFDs, SFILES/eSFILES, and learned flowsheet
representations encode increasing levels of process structure
\cite{theisen2023digitization,vogel2023sfiles,mann2024esfiles}.
However, these representations do not by themselves resolve the topology,
property-package, specification, and degrees-of-freedom decisions required for
executable equation-oriented simulation. Commercial simulators provide mature
modeling and automation capabilities, but typically do not expose the complete
symbolic model and execution state through a uniform, programmatically
inspectable substrate. In contrast, IDAES represents variables, constraints,
connections, specifications, and solver results as accessible Python objects.
This inspectability supports reproducible model assembly, structural validation,
and localized diagnosis. Prior IDAES studies consider
reinforcement-learning-based conceptual design
\cite{wang2023idaesrl} and knowledge-graph digital twins
\cite{zhang2026digitaltwins}, but do not investigate LLM-based multi-agent
systems for autonomous construction, validation, and repair of executable
equation-oriented process models. CRAFTS then builds on the code-native IDAES/Pyomo
ecosystem to integrate LLM agents with inspectable process-model objects and
solver states.

\textbf{Agentic process simulation.}
Recent systems apply LLMs and autonomous agents to automate simulation
planning \cite{tian2026textsimulation}, text-to-flowsheet generation
\cite{laub2026textflowsheet}, sketch-to-HYSYS conversion
\cite{bahamdan2026sketch2simulation}, Aspen Plus process development
\cite{yang2026ceproagents, 10.1021/acs.iecr.6c00963}, and
natural-language interfaces for AVEVA simulation, distillation design,
optimization, carbon accounting, microreactor design, and flowsheet correction
\cite{liang2026userfriendly,tan2026reasoningprocess,
pan2025chatmicroreactor,balhorn2024autocorrection}. Related studies further
explore domain adaptation, multi-agent optimization, scientific-agent
evaluation, and automated research workflows
\cite{zhang2026reactorfinetuning,zeng2025multiagentopt,
chen2025scienceagentbench,lu2026endtoend}.

Despite this progress, many existing approaches remain challenged by
(i) reliance on commercial simulation platforms with limited programmatic
access to internal model structures and execution states, which complicates
inspection, debugging, and reproducible repair;
(ii) dependence on hosted proprietary AI models, which restricts
local deployment, controllability, and practical adoption in industrial
settings; and (iii) insufficient integration of ChE knowledge into agentic flowsheet construction, resulting in weak consistency guarantees across topology, thermophysical properties, specifications, and simulator
capabilities. CRAFTS studies this combined setting on the open IDAES/Pyomo substrate, where role-specific agents propose engineering artifacts and deterministic gates control their promotion into executable models.

\section{Problem Setting}

\subsection{Task Definition}

Given a request bundle \(x_i=(r_i,p_i)\), curated ChE knowledge
(\(K\)), and verified IDAES-based simulator capabilities (\(C\)), CRAFTS generates
validated simulator-facing artifacts
\begin{equation}
A_i=f(x_i,K,C)=
\{G_i,V_i,T_i,S_i,B_i,E_i,O_i\},
\end{equation}
where \(r_i\) is the natural-language request and \(p_i\) is PFD evidence. The generated artifacts include a requirement interpretation artifact
(\(G_i\)), VisualGraphIR (\(V_i\)), TopologyIR (\(T_i\)), SpecIR (\(S_i\)), compiled BuildPlan (\(B_i\)), SolveReport \((E_i\)), and optional optimization artifact (\(O_i\)). \(K\) encodes reusable ChE knowledge for unit operations, thermodynamics, ports, DoF closure, initialization, and simulator constraints.

Following PSE practice that treats model structure and execution state as first-class objects \cite{biegler1997systematic,lee2021idaes}, we evaluate two complementary properties: structural reference alignment and executable workflow success. Structural reference is measured by macro $F1$ over normalized unit
records (UC), stream records (SC), and directed connection pairs (CC). For
request \(i\) and metric \(m\), let \(\widetilde{Y}_{i,m}\) and \(Y_{i,m}\)
denote predicted and reference sets:
\begin{equation}
F1_{i,m} =
\begin{cases}
0, & \text{no valid artifact},\\
1, & \widetilde{Y}_{i,m}=Y_{i,m}=\emptyset,\\
\dfrac{2|\widetilde{Y}_{i,m}\cap Y_{i,m}|}
{|\widetilde{Y}_{i,m}|+|Y_{i,m}|}, & \text{otherwise}.
\end{cases}
\label{eq:case-f1}
\end{equation}
The benchmark-level score is
\begin{equation}
\overline{F1}_m=\frac{1}{N}\sum_{i=1}^{N}F1_{i,m}.
\end{equation}

The Workflow Success (WS) measures whether a request completes all required stages,
including interpretation, topology validation, specification closure,
BuildPlan compilation, model construction, initialization, accepted solver
termination, and optional optimization. Let \(z_{i,s}\in\{0,1\}\) indicate
whether request \(i\) satisfies stage \(s\):
\begin{align}
W_i &= \prod_{s\in\mathcal{S}_i} z_{i,s},
\label{eq:workflow-case}\\
\mathrm{WS} &= \frac{100}{N}\sum_{i=1}^{N} W_i.
\label{eq:workflow-success}
\end{align}

UC, SC, and CC measure agreement with the frozen structural reference, whereas Workflow Success measures completion of the prescribed validation and execution contract; all metrics use the same fixed denominator. Explicit-request satisfaction and physical residuals are reported separately in the supplementary material.

\subsection{Artifact State}

CRAFTS maintains typed intermediate artifacts that preserve engineering
decisions throughout the model-construction process. The requirement artifact
\(G_i\) captures interpreted process objectives, constraints, ambiguities, and
provenance. When PFD evidence is available, VisualGraphIR \(V_i\) represents
visual entities and relations, while TopologyIR \(T_i\) and SpecIR \(S_i\)
encode the process structure and operating specifications required for simulator
construction.

A deterministic compiler transforms validated structural artifacts into an
executable construction plan:
\begin{equation}
\label{eq:buildplan}
B_i=\operatorname{compile}(T_i,S_i,C),
\end{equation}
where \(C\) specifies verified simulator capabilities. The resulting
SolveReport \(E_i\) records construction status, initialization results,
degrees-of-freedom closure, solver termination, physical checks, and failure
diagnostics. Optional optimization artifacts \(O_i\) finally store validated
optimization configurations and results.

These artifacts form the structured LangGraph state. Each role receives only the request, available evidence, curated ChE knowledge, promoted ancestors, and simulator capability constraints; evaluator references remain inaccessible during generation, repair, and execution.

\subsection{Promotion States}

Rather than relying on free-form agent outputs, CRAFTS promotes artifacts
through deterministic validation stages. Representation gates verify artifact
schema, entity consistency, and required relations. Compilation gates validate
unit availability, port compatibility, property-package support, graph
connectivity, and specification targets before model generation. Runtime gates
validate model construction, degrees-of-freedom closure, initialization,
solver termination, physical consistency, and optimization feasibility when
required.

A request is considered workflow-successful only when all required stages in
Eq.~\ref{eq:workflow-success} are satisfied. These promotion states provide
explicit checkpoints for diagnosis and bounded repair while preserving the
intermediate engineering state throughout execution.

\section{CRAFTS Workflow}

\subsection{Overview}

\begin{figure*}[t]
\centering
\includegraphics[
  width=1\textwidth,
  height=0.50\textheight,
  keepaspectratio
]{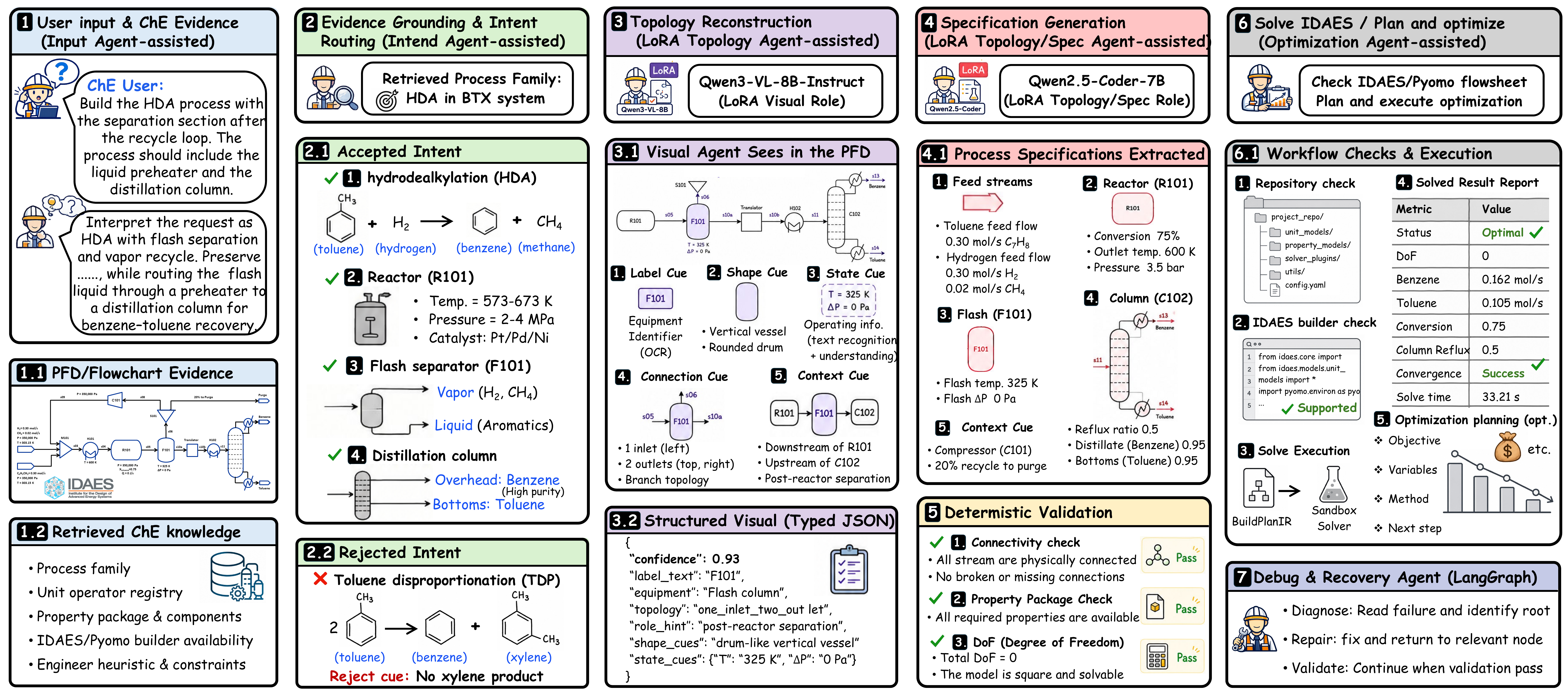}
\caption{CRAFTS artifact workflow illustrated through an HDA
(hydrodealkylation to benzene) example. Starting from the chemical engineer user request, PFD, and ChE-knowledge-informed evidence, the workflow performs intent routing, VisualGraphIR, TopologyIR and SpecIR generation, deterministic connectivity, property-package, and DoF, BuildPlan compilation, and IDAES/Pyomo execution before optional optimization. Failed artifacts are returned to their responsible stage for bounded diagnosis and repair through the LangGraph workflow architecture. LLM agents provide stage-specific engineering reasoning, while validation,
compilation, repair application, and numerical execution remain deterministic
workflow operations.}
\label{fig:architecture}
\end{figure*}

CRAFTS decomposes chemical-process model construction into staged agentic
reasoning and deterministic validation rather than a single code-generation task. Specialized agents progressively transform
engineering requests and available process evidence into typed intermediate
artifacts, including requirement interpretation, VisualGraphIR, TopologyIR,
SpecIR, BuildPlan, and SolveReport. Deterministic gates validate artifact
contracts, compile accepted representations, execute the resulting IDAES model,
and perform bounded stage-specific repairs. This separation allows LLM agents to
provide flexible engineering reasoning while preserving the reliability and
inspectability required for equation-oriented simulation. Figure
\ref{fig:architecture} illustrates the workflow on a toluene hydrodealkylation
(HDA) process, and Algorithm~\ref{alg:promotion} summarizes artifact promotion.

\begin{algorithm}[t]
\caption{ChE-informed artifact promotion}
\label{alg:promotion}
\begin{algorithmic}[1]
\STATE Interpret chemical engineer's natural-language requests/PFD cues into ChE-informed state \(G\)
\STATE Generate and validate VisualGraphIR \(V\)
\IF{visual contracts fail} \STATE bounded repair or fail closed \ENDIF
\STATE Assemble and validate TopologyIR \(T\)
\IF{topology contracts fail} \STATE bounded repair or fail closed \ENDIF
\STATE Generate SpecIR \(S\); validate schema, targets, and expected-DoF intent
\IF{specification contracts fail} \STATE bounded repair or fail closed \ENDIF
\STATE Compile accepted \(T\), \(S\), and \(C\) into hermetic BuildPlan \(B\)
\STATE Execute \(B\) in sandbox mode
\IF{the solved state passes execution checks and optimization is requested}
  \STATE validate variables/objective, release controls, and optimize
\ENDIF
\STATE Record reports, provenance, and terminal status \(E\)
\end{algorithmic}
\end{algorithm}

\subsection{Requirement and Structure Recovery}

The Input Understanding Agent extracts process objectives, operating
constraints, topology relations, and unresolved ambiguities from the original
request. The Intent Router maps these requirements to relevant chemical
engineering concepts and unit-operation relations using curated ChE knowledge.
The resulting requirement artifact provides a structured basis for downstream
model construction.

When PFD evidence is available, the Visual Agent converts
visual information into VisualGraphIR, representing process entities and their
relations. The Topology Agent then maps the interpreted requirements and visual
evidence to TopologyIR, defining the process structure required for simulation.
A deterministic topology validator checks schema validity, connectivity, port
compatibility, and simulator capability constraints before promotion.

\subsection{Specification Closure and Model Compilation}

After topology validation, the Specification Agent generates SpecIR to define
operating targets and close the degrees of freedom of the process model.
Deterministic specification gates verify target validity, variable resolution,
and consistency with the accepted topology before model construction.

The deterministic compiler then constructs the BuildPlan
\(B_i\), as defined in
Eq.~\ref{eq:buildplan}.
Unlike model-generation stages, compilation and execution are fully
deterministic to avoid LLM's flexible reasoning: promoted artifacts are translated into IDAES objects, initialized, and solved without further LLM intervention. The resulting SolveReport records numerical status, realized degrees of freedom, solver
termination, physical checks, and failure diagnostics.

\subsection{Diagnosis, Bounded Repair, and Optimization}

When validation or execution fails, CRAFTS localizes the failure to the
corresponding artifact stage rather than regenerating the entire model. A
debugging role analyzes the request, promoted artifacts, validator reports, and
solver diagnostics to identify the responsible layer, and a deterministic
failure router applies an allowlisted repair before returning the artifact to
its validation stage. Invalid artifacts are prevented from propagating
downstream, and runs that exhaust the repair budget fail closed. This
stage-specific repair strategy preserves intermediate engineering decisions and
supports transparent debugging of the model-construction process. After a valid solve, an optimization role then proposes an objective, decision variables, constraints, and target DoF. 

Promoted artifacts are maintained throughout the workflow, while validation, compilation, repair, simulation, and optimization remain non-LLM execution stages under fixed solver settings.

\section{Experiments}

\subsection{Benchmark}

OpenIDAES-450 contains 450 user-facing chemical-engineering requests paired
with executable IDAES process models, intermediate artifacts, and auditable
execution records. The benchmark covers unit-operation selection, topology
construction, property-package assignment, specification closure, initialization,
recycle handling, solver diagnosis, and optimization. We freeze 82 requests as
the held-out test set for all reported evaluations, while the remaining 368
requests are used only for training and development of CRAFTS-specific
components. Test membership, request order, and evaluation denominators remain fixed across all experiments. Cases derived from the same source model or process template are assigned to the same split, and duplicate requests and topology records are removed before evaluation.

\subsection{Evaluation Protocol}

All experiments use the same frozen 82-request test set, evaluator-only
references, structural scorers, and failure-retention rules. For framework
comparisons, we keep the CRAFTS role interfaces, artifact schemas,
deterministic gates, compiler, and execution settings fixed while replacing
only the orchestration framework. For model comparisons, alternative backends
are evaluated through the same role interfaces and workflow configuration.
Literature baselines retain their original architectures and are reported
separately. These comparisons evaluate complete system configurations under a shared artifact and execution protocol, rather than isolated foundation-model capabilities.

\subsection{Implementation Details}

CRAFTS is implemented as a LangGraph workflow with seven model-backed roles and deterministic validation and execution nodes. LangChain interfaces bind each role to its prompt, schema, structured-output parser, and artifact contract. The Visual Descriptor uses Qwen3-VL-8B-Instruct with supervised fine-tuning (SFT) via LoRA, while the Topology and Specification Agents use separate LoRA adapters on Qwen2.5-Coder-7B-Instruct; the remaining four roles use locally deployed Qwen3.6-27B with ChE-knowledge-informed prompting. All Qwen fine-tuning and local inference are performed on NVIDIA A40 GPUs, whereas external baseline models are accessed through their commercial APIs. Candidate outputs are archived and promoted only after passing deterministic fail-closed contracts. Full training settings, software versions, hardware details, and role-level configurations are provided in the supplementary material.

\section{Main Results}
\subsection{Overall comparison.}
Table~\ref{tab:full-workflow-450} compares architecture, orchestration, and
uniform-model controls on the frozen held-out requests. Workflow Success measures end-to-end completion under the shared CRAFTS/IDAES execution protocol, while UC, SC, and CC independently measure structural agreement with evaluator-only references. A request contributes zero only when it
produces no scorable topology artifact, and all 82 requests remain in every
denominator. CRAFTS succeeds on 75 of 82 requests (91.5\%), with a
two-sided 95\% Wilson interval of 83.4--95.8\%. All methods are evaluated on the same 82 requests using fixed decoding and execution settings, and the reported results are point estimates on this frozen test split.

\begin{table*}[t]
\centering
\setlength{\tabcolsep}{5pt}
\renewcommand{\arraystretch}{1.06}

\begin{tabular*}{\textwidth}{
    @{\extracolsep{\fill}}
    P{0.4\textwidth}
    cccc
    @{}
}
\toprule
\textbf{Method}
& \multicolumn{1}{c}{\textbf{Execution}}
& \multicolumn{3}{c}{\textbf{Structural F1}} \\
\cmidrule(lr){2-2}
\cmidrule(lr){3-5}
& Workflow Success (\%) $\uparrow$
& UC $\uparrow$
& SC $\uparrow$
& CC $\uparrow$ \\
\midrule

\multicolumn{5}{@{}l}{\textbf{Architecture controls }} \\
CeProAgents-style~\cite{yang2026ceproagents}
& 74.4 & 0.486 & 0.188 & 0.300 \\
Text-to-Simulation-style~\cite{tian2026textsimulation}
& 68.3 & 0.456 & 0.210 & 0.293 \\
Sketch2Simulation-style~\cite{bahamdan2026sketch2simulation}
& 76.8 & 0.454 & 0.196 & 0.289 \\
\midrule

\multicolumn{5}{@{}l}{\textbf{Orchestration-framework controls}} \\
OpenAI Agents SDK ~\cite{openaiAgentsSdk2026}
& 79.3 & 0.571 & 0.363 & 0.474 \\
Microsoft Agent Framework~\cite{microsoftAgentFramework2026}
& 72.0 & 0.544 & 0.364 & 0.453 \\
CrewAI~\cite{crewaiDocs2026}
& 76.8 & 0.556 & 0.360 & 0.466 \\

\midrule
\multicolumn{5}{@{}l}{\textbf{Uniform-model controls}} \\
GPT-5 mini~\cite{openaiGpt5SystemCard2025}
& 54.9 & 0.392 & 0.280 & 0.203 \\
Claude Sonnet 4.6~\cite{anthropicClaudeSonnet46SystemCard2026}
& 45.1 & 0.210 & 0.159 & 0.167 \\
Gemini 3.5 Flash~\cite{googleDeepMindGemini35Flash2026}
& 28.0 & 0.215 & 0.166 & 0.128 \\

\midrule
\textbf{CRAFTS (role-adaptive local Qwen)}
& \best{91.5}
& \best{0.815}
& \best{0.791}
& \best{0.782} \\

\bottomrule
\end{tabular*}

\caption{
Frozen 82-request results (higher is better). Workflow Success is the
percentage of requests that pass every required stage in
Eq.~\ref{eq:workflow-success} and is reported independently of
fixed-denominator structural macro F1. Both the architecture and
orchestration-framework controls use GPT-5 mini with the shared IDAES
execution substrate and structured-output schema; the architecture controls
are author reimplementations of the cited agent decompositions.
}
\label{tab:full-workflow-450}
\end{table*}

\textbf{Observed differences and relational assembly.}
CRAFTS obtains the highest observed value in every column. Relative to the
strongest non-CRAFTS value in each column, the measured differences are
12.2 percentage points for Workflow Success and 0.244/0.427/0.308 for
UC/SC/CC. The larger SC and CC gains are particularly important in chemical
process modeling: identifying individual units such as a flash, reactor, or
compressor is insufficient unless material identity is preserved across
branches, stream connectivity remains consistent, and every connection
terminates at a legal, directionally correct port. CRAFTS therefore improves
not only equipment recovery but also relational flowsheet assembly, which
directly determines whether an equation-oriented model can be constructed and
solved. UC/SC/CC measure structural reference alignment, whereas Workflow
Success additionally incorporates topology, thermodynamic, specification,
construction, initialization, numerical, and requested-optimization gates.

\textbf{Role-adapted local models.}
Under the same role graph and evaluation protocol, the locally deployed
role-adaptive Qwen configuration outperforms all evaluated uniform
commercial-backend substitutions. Its Workflow Success exceeds GPT-5 mini,
Claude Sonnet 4.6, and Gemini 3.5 Flash by 36.6, 46.4, and 63.5 percentage
points, respectively. These system-level results show that the role-adaptive configuration is more effective than the evaluated uniform-model substitutions under the same workflow.

The visual stage requires multimodal interpretation of process symbols,
spatial relations, and stream connectivity; topology construction requires
schema-constrained graph generation and simulator compatibility reasoning; and
specification closure requires precise mapping between engineering targets,
variables, and degrees of freedom. The gains from role adaptation therefore
reflect capability alignment between agents and engineering subtasks rather
than a ranking of foundation models themselves.

\begin{figure}[t]
\centering
\includegraphics[width=0.4\textwidth]{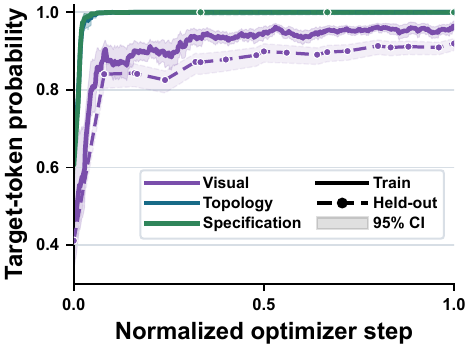}
\caption{LoRA SFT target-token probability over normalized optimizer steps for
  the Visual Descriptor (Qwen3-VL-8B) and the Topology and Specification Agents
  (Qwen2.5-Coder-7B). Solid and dashed curves denote training and held-out
  validation, respectively; shading indicates 95\% confidence intervals.}
\label{fig:lora-sft-probability}
\end{figure}

The results support the combined heterogeneous-backbone and role-specific
adaptation design over uniformly assigning one hosted model to all roles.
Figure~\ref{fig:lora-sft-probability} shows that the topology and specification
adapters converge faster than the visual adapter under the reported SFT metric,
consistent with their more structured output spaces.

\subsection{Ablation Study}

Table~\ref{tab:ablation-450} evaluates five system variants on the same frozen
82-request split. The untuned condition removes the three role-specific LoRA
adapters; the ChE and visual conditions remove the corresponding model-visible
evidence; joint topology--specification generation removes the independent
TopologyIR-to-SpecIR promotion boundary; and the no-Debug condition disables
bounded repair.

\begin{table}[t]
\centering
\setlength{\tabcolsep}{1pt}
\renewcommand{\arraystretch}{1.06}
\begin{tabular*}{\columnwidth}{@{\extracolsep{\fill}}P{0.34\columnwidth}cccc@{}}
\toprule
\textbf{Ablated condition} & \multicolumn{1}{c}{\textbf{Execution}} &
\multicolumn{3}{c}{\textbf{Structural F1}} \\
\cmidrule(lr){2-2}\cmidrule(lr){3-5}
& Work.\ Success (\%) $\uparrow$ &
UC $\uparrow$ & SC $\uparrow$ & CC $\uparrow$ \\
\midrule
Local Qwen (no FT) & 36.6 & 0.242 & 0.109 & 0.135 \\
No ChE knowledge & 0.00 & 0.058 & 0.116 & 0.083 \\
No visual evidence & 25.6 & 0.174 & 0.078 & 0.105 \\
Merged topo.--spec. & 30.5 & 0.273 & 0.117 & 0.133 \\
No Debug replay & 18.3 & 0.151 & 0.104 & 0.090 \\
\midrule
\textbf{Full CRAFTS} &
\best{91.5} & \best{0.815} & \best{0.791} & \best{0.782} \\
\bottomrule
\end{tabular*}
\caption{Ablation study of CRAFTS on the frozen 82-request test split. The
results quantify the impact of role-specific adaptation, ChE-informed
reasoning, visual evidence, topology--specification separation, and
debugging-based repair on executable process-model construction.}
\label{tab:ablation-450}
\end{table}

\textbf{Ablation observations.}
Removing role-specific adaptation reduces Workflow Success from 91.5\% to
36.6\%, with larger losses in SC and CC than in UC, showing that stream
identity and directed port binding are more demanding than equipment
recognition. Removing model-visible ChE knowledge leaves limited local
structural overlap but yields no workflow-successful request, because unit,
component, phase, property-package, specification, and initialization choices
must remain jointly compatible. Removing PFD evidence produces the largest SC
drop, from 0.791 to 0.078, reflecting the role of arrows, branches, merges, and
recycle loops in stream reconstruction. Joint topology--specification
generation reduces Workflow Success to 30.5\%, supporting independent
TopologyIR promotion before numerical specification. Disabling Debug replay
reduces Workflow Success to 18.3\%, showing that bounded stage-specific repair
is integral to the complete workflow. 

This observation is consistent with the audit that 75 of 82 cases required at
least one debugging interaction, demonstrating that iterative diagnosis is
intrinsic to practical process-model construction.

\section{Discussion and Limitations}

\textbf{Engineering insight.}
CRAFTS reflects a central principle of chemical-process modeling: executable
flowsheets emerge from coordinated engineering stages rather than a single
generation step. The larger improvements in SC and CC than in UC indicate that
the primary challenge is not recognizing individual equipment types, but
preserving relational process structure. In equation-oriented simulation,
valid models require consistent stream identity, directional port connectivity,
compatible property domains, and numerically meaningful specifications across
the entire flowsheet. These results suggest that process-model intelligence
depends strongly on maintaining structured engineering relations rather than
only recovering isolated process entities.

\textbf{Why artifact-based agents work.}
The effectiveness of CRAFTS arises from separating flexible LLM-based reasoning
from deterministic engineering execution. Specialized agents handle
stage-specific decisions, while typed artifacts preserve explicit engineering
states across topology construction, specification closure, diagnosis, and
execution. Unlike free-form agent communication, these artifacts provide a
shared and inspectable state that allows deterministic gates to validate
structural, physical, and numerical consistency before information propagates
downstream. Bounded repair then revises the responsible modeling layer using
validated intermediate states rather than regenerating the entire model. This
design matches the iterative workflow used by chemical engineers, where
flowsheet structure, numerical closure, and solver behavior are progressively
refined through diagnosis and correction.

\textbf{Limitations and future extensions.}
The current evaluation focuses on ChE modeling tasks within the verified IDAES
capability surface represented by OpenIDAES-450. Extending CRAFTS to broader
process families, additional property models, and more diverse optimization
settings remains an important direction. The current benchmark also emphasizes
auditable process-model construction; future work can investigate larger-scale
industrial flowsheets, richer multimodal inputs, and adaptive knowledge
resources. The artifact and execution traces introduced by CRAFTS provide a
foundation for such extensions.

\section{Conclusion}

We presented CRAFTS, a locally deployable LLM multi-agent system for
constructing executable chemical-process models from underspecified requests.
Through seven role-adaptive agents and deterministic workflow nodes, CRAFTS
transforms natural-language and PFD evidence into typed VisualGraphIR,
TopologyIR, SpecIR, BuildPlan, SolveReport, and optional optimization
artifacts, with engineering gates enforcing structural, thermodynamic,
numerical, and execution contracts.

CRAFTS targets equation- and code-oriented chemical-process modeling over the
open IDAES/Pyomo substrate, exposing process structure, equations, constraints,
DoF, solver diagnostics, and optimization controls as inspectable program
objects. On the frozen 82-request OpenIDAES-450 split, CRAFTS achieves 91.5\% Workflow Success and UC/SC/CC macro F1 scores of 0.815/0.791/0.782, obtaining the best observed results among the evaluated architecture, orchestration, and uniform-model controls under the shared workflow. These results demonstrate the potential of ChE-informed, locally deployable LLM collaboration as an effective and auditable approach for executable chemical-process model construction.
\clearpage
\appendix
\raggedbottom
\section{Role-Specific Fine-Tuning Details}
\label{sec:finetuning-appendix}

Only the Visual Descriptor, Topology Agent, and Specification Agent use
role-specific LoRA adapters; the other four roles use ordinary Qwen. The frozen
82-case benchmark is excluded from SFT, prompt construction, and the ChE
knowledge base. The complementary 368 cases supply the training evidence and
transferable engineering guidance. Controlled transformations yield
5,000/1,000/1,000 train, validation, and test records for each text adapter and
1,033/238/240 multimodal records for the visual adapter. These are role-level
records rather than additional benchmark cases.

\begin{table}[htbp]
\centering
\small
\setlength{\tabcolsep}{3.5pt}
\begin{tabular*}{\textwidth}{@{\extracolsep{\fill}}P{0.15\textwidth}P{0.20\textwidth}P{0.41\textwidth}cc@{}}
\toprule
Fine-tuned role & Backbone and adapter & Supervised contract & Epochs & Steps \\
\midrule
Visual Descriptor & Qwen3-VL-8B-Instruct + LoRA &
Raster PFD and role prompt to validated VisualGraphIR with equipment, labels,
directions, branches, and recycle cues & 3 & 732 \\
Topology Agent & Qwen2.5-Coder-7B-Instruct + LoRA &
Normalized request, typed visual evidence, and ChE guidance to candidate
TopologyIR with units, ports, arcs, packages, and terminals & 2 & 1,250 \\
Specification Agent & Qwen2.5-Coder-7B-Instruct + LoRA &
Accepted TopologyIR and numerical evidence to SpecIR with targets, values,
units, provenance, fix intent, and solve intent & 2 & 1,250 \\
\bottomrule
\end{tabular*}
\caption{Role-specific fine-tuning configuration. Adapters are optimized
independently because the three roles expose different schemas and validation
surfaces. Steps are optimizer steps from the frozen training traces.}
\label{tab:arxiv-finetune-setup}
\end{table}

\subsection{Software, Hardware, and Promotion}

The frozen jobs use Python 3.10.20, PyTorch 2.6.0 with CUDA 12.4,
Transformers 5.7.0, PEFT 0.19.1, Tokenizers 0.22.2, and Safetensors 0.7.0.
The Visual Descriptor additionally uses bitsandbytes 0.49.2 for four-bit NF4
loading, Accelerate 1.13.0 for device placement, Pillow 12.2.0, and NumPy
2.2.6. Training and local inference run on NVIDIA A40 GPUs. PEFT 0.19.1 is
recorded in every adapter; the remaining pins were reconstructed from the
persistent interpreter on 2026-07-22 because the original runs did not retain a
complete transitive lock file.

Checkpoint promotion follows the artifact contract of each role rather than
training loss alone. VisualGraphIR, TopologyIR, and SpecIR checkpoints must pass
their respective schema and engineering gates. Figure 3 reports held-out SFT
target-token probability over normalized optimizer steps; solid curves
aggregate the multi-seed experiments and shaded regions show their 95\%
confidence intervals.

\clearpage
\section{Web Interface and Audited Case Demonstrations}
\label{sec:case-demos}

\subsection{Web Interface Demonstration}

\begin{figure}[htbp]
\centering
\includegraphics[width=0.96\textwidth]{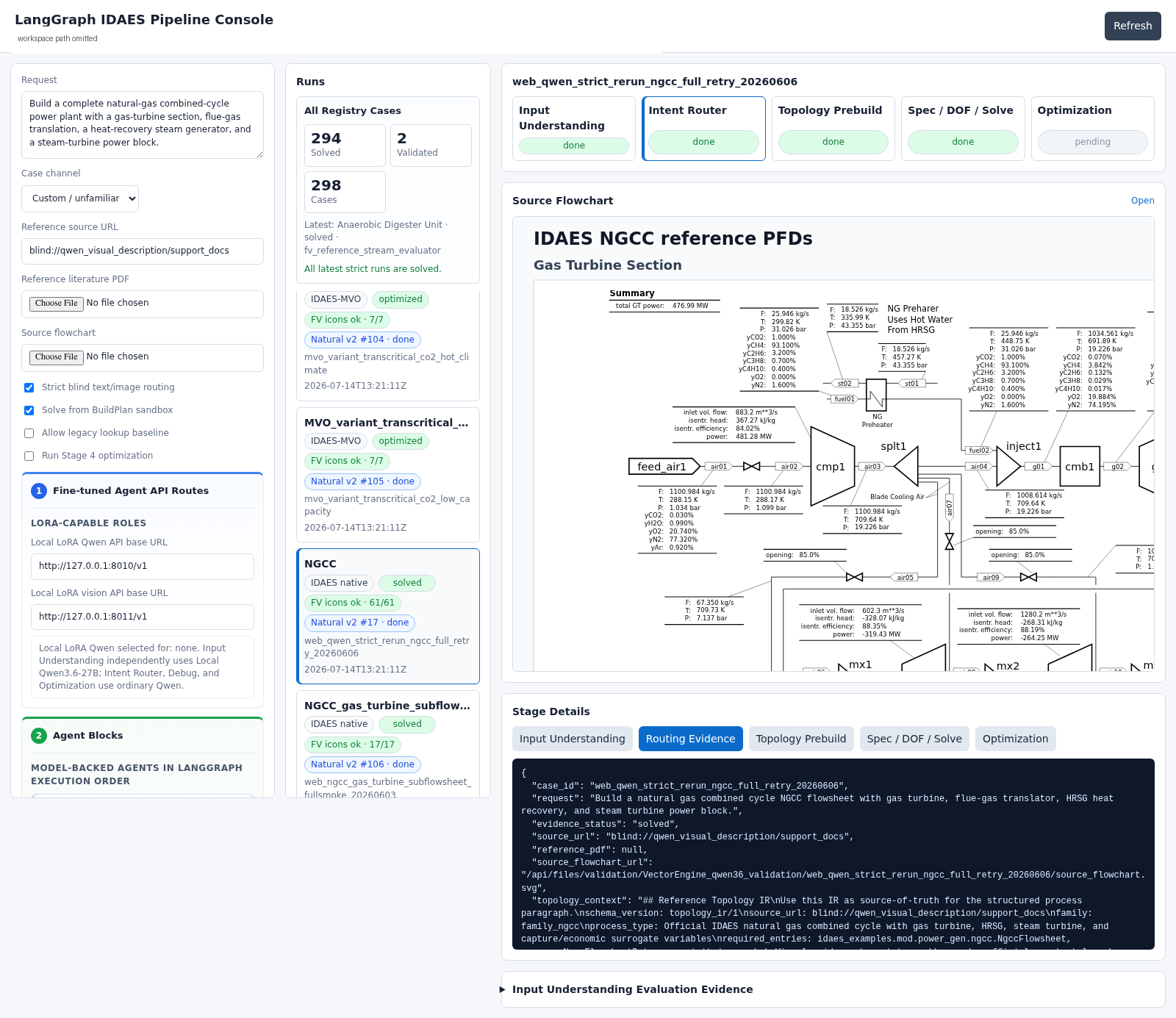}
\caption{Web interface demonstration for a saved NGCC trajectory. Request
controls and the run list appear at left; stage status, source PFD, and selected
artifact details appear at right. The screenshot documents the interaction and
audit surfaces only; benchmark results use the frozen artifacts described in
the main paper.}
\label{fig:web-interface-demo}
\end{figure}

Figure~\ref{fig:web-interface-demo} shows the local browser interface used to
inspect a saved NGCC trajectory. The interface exposes the user request and
inputs, saved runs, ordered stage status, source PFD, and the sanitized artifact
selected for inspection. It reads persisted run records and does not modify
model-visible context, execution, or scoring.

\clearpage
\subsection{Audited Case Demonstrations}

The following five demonstrations reproduce frozen, artifact-linked
trajectories from the held-out evaluation. Each trace preserves the request,
role handoffs, typed artifacts, deterministic gates, execution report, and
Debug state from one run. They illustrate system behavior and do not change the
fixed-denominator results reported in the main text.

\subsubsection{HDA Flash}
\begin{tracecasebox}{HDA flash: complete interaction trace}
\scriptsize
\runbar{}

\tracetool{LangGraph collaboration profile}{Audit report}{Input Understanding,
Intent, Visual, Topology, Specification, deterministic validation, initialization,
solve, and reporting form one artifact-consistent trajectory. Text roles use
medium reasoning and the Visual Descriptor uses low reasoning. Five scoped
model-agent calls generate 15 visibility-audit events, comprising five each for prepared
calls, responses, and finalized calls, with zero boundary errors. The Debug
supervisor is armed before Stage 1. All five role outputs are accepted on
attempt 0 with no schema repair, fallback, or Debug replay; the deterministic
solve reaches the report boundary.}

\tracestage{1}{Input Understanding: preserve requirements, relations, and negations}

\traceuser{Root User}{Input Understanding}{Build an HDA hydrodealkylation
flowsheet with feed mixing, heating, reaction, flash separation, purge,
compression, and vapor recycle. Use the compact flash configuration without a
downstream distillation column.}

\traceagent{Input Understanding}{Intent Router}{Normalize the task as
\code{build\_and\_solve} in the HDA hydrodealkylation domain. Required features
are \code{feed\_mixing}, \code{heater}, \code{reactor},
\code{flash\_separator}, \code{purge}, \code{compressor},
\code{vapor\_recycle}, and \code{compact\_flash\_configuration}; the negated
feature is \code{distillation\_column}. Record the six directed relations:
feed mixing feeds the heater, the heater feeds the reactor, the reactor feeds
the flash, the flash vapor feeds both compressor and purge, and the compressor
recycles to feed mixing. Record the bounded assumption that standard HDA
connections fill details absent from the request, no unresolved ambiguity,
\code{needs\_clarification=false}, and confidence 0.95. The first structured
role output passes the contract.}

\tracestage{2}{Intent: recover the ChE process boundary and downstream obligations}

\traceagent{Intent Router}{ChE evidence / Visual Descriptor}{Interpret the
request as \code{HDA Flash} in the \code{reaction\_separation} family with
confidence 1.0 while preserving the explicit no-distillation boundary. The ChE
interpretation requires feed mixing, reaction preheating, hydrodealkylation,
vapor--liquid separation, a purge, recompression, vapor recycle, and a separate
benzene-rich liquid path. The first structured role output passes the contract.}

\tracestage{3}{Visual evidence: translate the PFD into an inspectable observation}

\tracetool{ChE evidence coordinator}{Visual Descriptor}{Provide the process
request together with authoritative HDA process documentation as a role-scoped
engineering evidence packet. It covers visible equipment labels, stream
directions, the vapor-recycle cue, the purge branch, the liquid-product path,
and operating annotations. Only the Visual Descriptor receives the raster and
its role-specific ChE guidance; the image is process evidence, not an evaluator
topology or executable implementation.}

\traceagent{Visual Descriptor}{VisualGraphIR Validator / Topology Agent}{Act as
a modeling-assistance perception adapter upstream of topology synthesis. Convert
the raster PFD into a typed visual record of visible equipment, label text,
arrow direction, products, branches, recycle cues, confidence, evidence links,
and uncertainty. Do not independently choose IDAES property packages, ports,
specifications, or executable model code. The Topology Agent will combine this
validated observation with routed ChE evidence and simulator capability
constraints. The promoted artifact contains 12 visual nodes and 12 directed
edges: nodes \code{I101,I102,M101,H101,R101,F101,S101,C101,F102,P101,P102,P103}
and edges \code{I101->M101}, \code{I102->M101}, \code{M101->H101},
\code{H101->R101}, \code{R101->F101}, \code{F101->S101},
\code{S101->P103}, \code{S101->C101}, \code{C101->M101} (recycle),
\code{F101->F102}, \code{F102->P101}, and \code{F102->P102}. It also reads the
two feed conditions, heater/reactor/flash labels, 20\% purge, compressor
pressure, second-flash conditions, and benzene, toluene, and purge products.
There are no uncertainty entries. A manuscript rendering of the actual JSON is:
\begin{tracejsonbox}
\{\\
\ \ "schema\_version": "visual\_graph\_ir/2",\\
\ \ "nodes": [\\
\ \ \ \ \{"visual\_id":"M101", "label\_text":"M101",\\
\ \ \ \ \ "equipment\_class\_guess":"Mixer", "bbox":[0.18,0.4,0.3,0.6],\\
\ \ \ \ \ "confidence\_0\_1":1.0,\\
\ \ \ \ \ "evidence\_ref":"Triangular mixer icon with label M101"\},\\
\ \ ],\\
\ \ "edges": [\\
\ \ \ \ \{"visual\_edge\_id":"e09", "source\_visual\_id":"C101",\\
\ \ \ \ \ "destination\_visual\_id":"M101", "label\_text":"",\\
\ \ \ \ \ "direction\_confidence\_0\_1":1.0, "is\_recycle\_guess":true,\\
\ \ \ \ \ "evidence\_ref":"Arrow from C101 back to M101"\}\\
\ \ ],\\
\ \ "uncertainties": []\\
\}
\end{tracejsonbox}
The full JSON additionally retains bounding boxes, all feed/product boundary
nodes, every evidence reference, and the remaining ten directed edges.}

\tracetool{VisualGraphIR Validator}{Topology Agent}{Validate schema, unique
visual identifiers, normalized bounding boxes, edge endpoints, and evidence
boundary. Accept the first Visual Descriptor output with zero
canonicalizations and no text-only fallback.}

\tracestage{4}{Topology: combine observations with ChE and simulator constraints}

\traceagent{Topology Agent}{Topology Validator / Prebuilder}{Fuse the typed
visual record with HDA ChE constraints and emit seven IDAES unit records, all
on \code{BTHM\_params}: \code{M101} Mixer (toluene feed, hydrogen feed, and
vapor recycle), \code{H101} Heater (reactor preheater), \code{R101} CSTR (HDA
reactor with \code{reaction\_params}), \code{F101} Flash (noncondensable/liquid
split), \code{S101} Splitter (purge and recycle), \code{C101} PressureChanger
(isothermal recycle compressor), and \code{F102} Flash (benzene-rich product
flash). Emit the seven exact stream endpoints:
\code{s03 M101.outlet->H101.inlet};
\code{s04 H101.outlet->R101.inlet};
\code{s05 R101.outlet->F101.inlet};
\code{s06 F101.vap\_outlet->S101.inlet};
\code{s08 S101.recycle->C101.inlet};
\code{s09 C101.outlet->M101.vapor\_recycle}; and
\code{s10 F101.liq\_outlet->F102.inlet}. The first typed output passes without
fallback.}

\tracestage{5}{Specification: close operating choices and degrees of freedom}

\tracetool{Topology Validator / Prebuilder}{Specification Agent}{Check unit and
package surfaces, endpoints, ports, terminal streams, and construction
compatibility. The promoted TopologyIR passes with exactly 7 units, 7 arcs, and
zero errors.}

\traceagent{Specification Agent}{DOF Checker}{Emit all 16 grounded SpecIR
entries one-to-one. The provenance marker \emph{ChE-agent/DoF-validated}
denotes a Specification Agent proposal accepted by the deterministic DoF gate.
\textbf{Unit configuration (5):}
\code{R101.\allowbreak{}conversion} = \code{0.\allowbreak{}75} dimensionless
(toluene conversion target; ChE-agent/DoF-validated);
\code{R101.\allowbreak{}heat\_\allowbreak{}duty[\allowbreak{}0]\allowbreak{}} =
\code{0} W (adiabatic reactor duty; ChE-agent/DoF-validated);
\code{F101.\allowbreak{}deltaP[\allowbreak{}0]\allowbreak{}} = \code{0} Pa
(first flash pressure drop; ChE-agent/DoF-validated);
\code{S101.\allowbreak{}split\_\allowbreak{}fraction[\allowbreak{}0,
'purge']\allowbreak{}} = \code{0.\allowbreak{}2} dimensionless (purge split
fraction; ChE-agent/DoF-validated); and
\code{F102.\allowbreak{}deltaP[\allowbreak{}0]\allowbreak{}} =
\code{-\allowbreak{}200000} Pa (second flash pressure drop;
ChE-agent/DoF-validated). \textbf{Stream configuration (11):}
\code{M101.\allowbreak{}toluene\_\allowbreak{}feed.\allowbreak{}flow\_\allowbreak{}mol\_\allowbreak{}phase\_\allowbreak{}comp[\allowbreak{}0,
'Liq', 'toluene']\allowbreak{}} = \code{0.\allowbreak{}3} mol/s (toluene feed
molar flow; ChE-agent/DoF-validated);
\code{M101.\allowbreak{}hydrogen\_\allowbreak{}feed.\allowbreak{}flow\_\allowbreak{}mol\_\allowbreak{}phase\_\allowbreak{}comp[\allowbreak{}0,
'Vap', 'hydrogen']\allowbreak{}} = \code{0.\allowbreak{}3} mol/s (hydrogen feed
molar flow; ChE-agent/DoF-validated);
\code{M101.\allowbreak{}hydrogen\_\allowbreak{}feed.\allowbreak{}flow\_\allowbreak{}mol\_\allowbreak{}phase\_\allowbreak{}comp[\allowbreak{}0,
'Vap', 'methane']\allowbreak{}} = \code{0.\allowbreak{}02} mol/s (methane
impurity in hydrogen feed; ChE-agent/DoF-validated);
\code{M101.\allowbreak{}toluene\_\allowbreak{}feed.\allowbreak{}temperature[\allowbreak{}0]\allowbreak{}} =
\code{303.\allowbreak{}2} K (toluene feed temperature; ChE-agent/DoF-validated);
\code{M101.\allowbreak{}hydrogen\_\allowbreak{}feed.\allowbreak{}temperature[\allowbreak{}0]\allowbreak{}} =
\code{303.\allowbreak{}2} K (hydrogen feed temperature; ChE-agent/DoF-validated);
\code{M101.\allowbreak{}toluene\_\allowbreak{}feed.\allowbreak{}pressure[\allowbreak{}0]\allowbreak{}} =
\code{350000} Pa (toluene feed pressure; ChE-agent/DoF-validated);
\code{M101.\allowbreak{}hydrogen\_\allowbreak{}feed.\allowbreak{}pressure[\allowbreak{}0]\allowbreak{}} =
\code{350000} Pa (hydrogen feed pressure; ChE-agent/DoF-validated);
\code{H101.\allowbreak{}outlet.\allowbreak{}temperature[\allowbreak{}0]\allowbreak{}} =
\code{600} K (reactor preheat target; ChE-agent/DoF-validated);
\code{F101.\allowbreak{}vap\_\allowbreak{}outlet.\allowbreak{}temperature[\allowbreak{}0]\allowbreak{}} =
\code{325} K (first flash vapor temperature; ChE-agent/DoF-validated);
\code{C101.\allowbreak{}outlet.\allowbreak{}pressure[\allowbreak{}0]\allowbreak{}} =
\code{350000} Pa (recycle compressor outlet pressure;
ChE-agent/DoF-validated); and
\code{F102.\allowbreak{}vap\_\allowbreak{}outlet.\allowbreak{}temperature[\allowbreak{}0]\allowbreak{}} =
\code{375} K (second flash vapor temperature; ChE-agent/DoF-validated). No
assignment is replaced by a count; the first structured output passes without
fallback.}

\tracetool{DOF Checker}{BuildPlan Compiler}{Pass the deterministic consistency
gate with zero errors and zero warnings, 16 specifications, and expected
steady-state DoF 0.}

\tracestage{6}{Build and initialize the accepted white-box model}

\tracetool{BuildPlan Compiler}{Initialization / Native Solver}{Compile the
promoted TopologyIR and SpecIR into a hermetic JSON BuildPlan. Retain the
seven units, seven arcs, benzene/toluene/hydrogen/methane components in liquid
and vapor phases, the HDA property/reaction-package surface, terminal feeds
\code{M101.toluene\_feed} and \code{M101.hydrogen\_feed}, and terminal products
\code{S101.purge}, \code{F102.vap\_outlet}, and \code{F102.liq\_outlet}.
Declare SequentialDecomposition in the order
\code{M101,H101,R101,F101,S101,C101,F102}, with \code{H101.inlet} as the
recycle tear destination.}

\tracestage{7}{Solve, inspect diagnostics, repair, and optimize only when eligible}

\tracetool{Native Solver}{Failure Router / Report}{Complete the IDAES HDA solve
with \code{pass=true}, termination \code{optimal}, final DoF 0, no error, and
12 persisted stream records. F102 vapor and liquid products are both 375 K and
approximately 150 kPa. The liquid product contains 0.06262 mol/s liquid benzene
and 0.03226 mol/s liquid toluene; the vapor product contains 0.14198 mol/s vapor
benzene and 0.03026 mol/s vapor toluene. The purge is 325 K and approximately
350 kPa.}

\traceagent{Debug Supervisor audit (standby)}{Failure Router / Report}{No fault
is routed to Debug, so no diagnosis, repair artifact, or replay is generated.
The supervisor remains armed from before Stage 1 through terminal acceptance.}

\tracetool{Failure Router}{Optimization Agent / Report}{Accept the passing
TopologyIR, SpecIR, DoF, and native-solve reports, close the bounded repair
branch, and pass the accepted simulator state to Optimization only when the
request makes that role eligible.}

\traceagent{Optimization Agent}{Deterministic runner / Report}{Optimization is
not invoked because the current request is build-and-solve only; persist the
not-invoked state and make no design-optimum claim.}

\tracetool{IDAES FV exporter}{Web / manuscript report}{After solve and any
eligible optimization, export the same-run flowsheet visualization with 22 FV
cells. Retain the ChE process evidence, accepted TopologyIR sketch, FV graph,
and terminal numerical evidence for this same LangGraph trajectory.}
\end{tracecasebox}

\casefigure{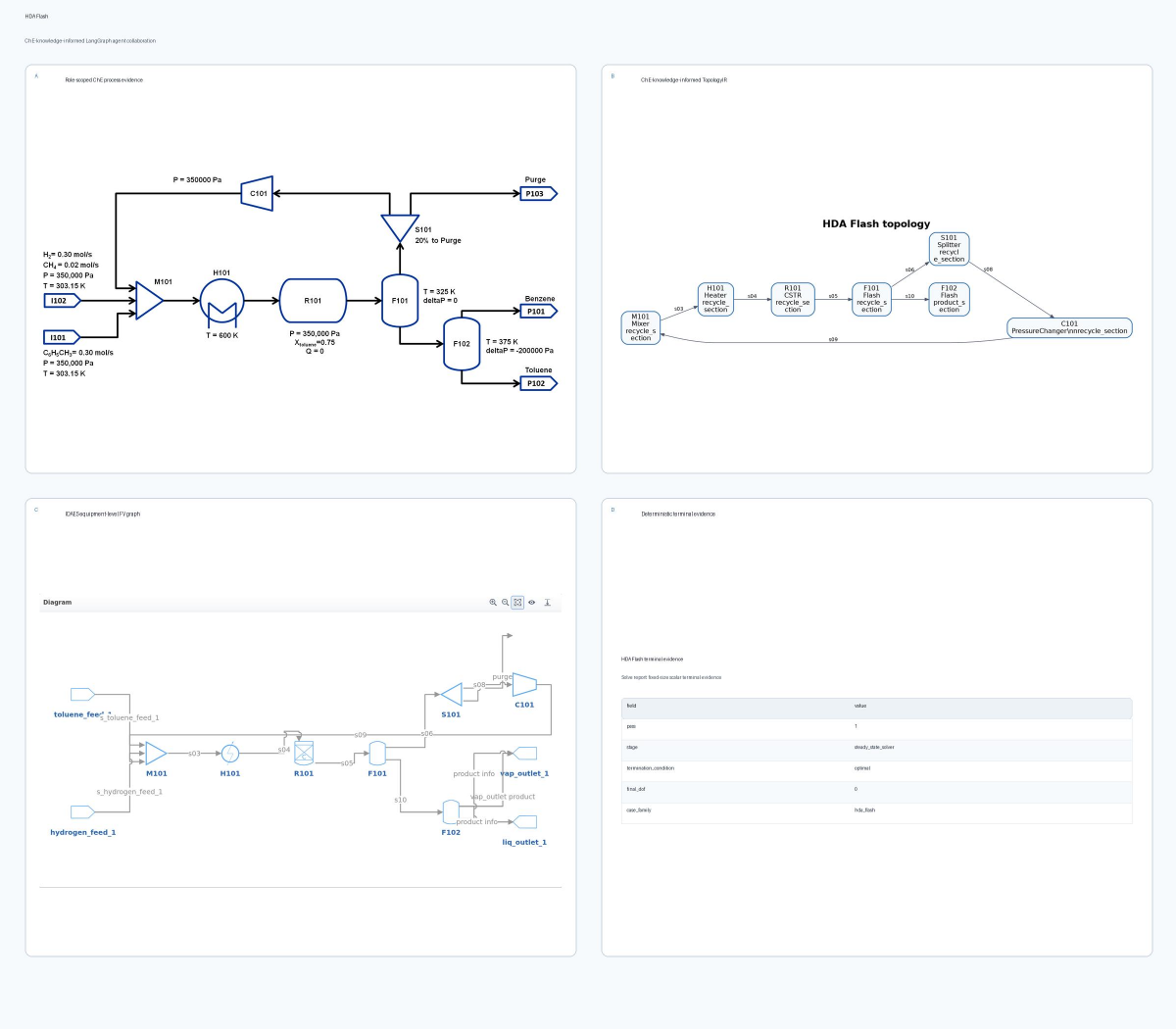}
{HDA flash collaboration audit bundle. The compact flash and recycle boundary
is kept separate from HDA distillation.}
{fig:arxiv-case-hda-flash}

\subsubsection{HDA Distillation}
\begin{tracecasebox}{HDA distillation---complete interaction trace}
\scriptsize
\runbar{}

\tracetool{LangGraph collaboration profile}{Case report}{Input Understanding, Intent,
Visual, Topology, Specification, deterministic validation, execution, and conditional
Debug/Optimization roles form one artifact-consistent trajectory. Role reasoning is bounded by
medium. The visibility gateway records 5 scoped model-agent
calls and 15 visibility-audit events with zero boundary errors.
Structured-output repairs=0; Debug replays=0. The
Debug supervisor is armed before Stage 1. The deterministic execution returns \code{solved} with
final acceptance at the report boundary.}

\tracestage{1}{Input Understanding: preserve requirements, relations, and negations}

\traceuser{Root User}{Input Understanding}{Build an HDA hydrodealkylation flowsheet with feed mixing, heating, reaction, flash separation, purge, compression, and vapor recycle. Route the flash liquid through the transfer and preheat section to a distillation column for benzene and toluene recovery.}

\traceagent{Input Understanding}{Intent Router}{Normalize the request as
\code{build\_\allowbreak{}and\_\allowbreak{}solve} in HDA hydrodealkylation. Required
features: \code{feed\_\allowbreak{}mixing}; \code{heater}; \code{reactor}; \code{flash\_\allowbreak{}separator}; \code{purge}; \code{compressor}; \code{vapor\_\allowbreak{}recycle}; \code{liquid\_\allowbreak{}preheater}; \code{distillation\allowbreak{}\_\allowbreak{}column}. Negated features:
none. Directed relations: feed\_mixing feeds heater; heater feeds reactor; reactor feeds flash\_separator; flash\_separator vapor\_outlet\_feeds compressor; compressor recycles\_to feed\_mixing; flash\_separator liquid\_outlet\_feeds liquid\_preheater; liquid\_preheater feeds distillation\_column.
Assumptions: Standard HDA topology assumed: reactor effluent goes to flash; vapor is compressed and recycled; liquid goes to distillation.;
ambiguities: none;
clarification required=\code{False}; confidence
0.95.}

\tracestage{2}{Intent: recover the ChE process boundary and downstream obligations}

\traceagent{Intent Router}{ChE evidence coordinator / Visual Descriptor}{Interpret the
request as HDA Distillation in the distillation domain.
Route the required equipment and process features---\code{feed\_\allowbreak{}mixing}; \code{heater}; \code{reactor}; \code{flash\_\allowbreak{}separator}; \code{purge}; \code{compressor}; \code{vapor\_\allowbreak{}recycle}; \code{liquid\_\allowbreak{}preheater}; \code{distillation\allowbreak{}\_\allowbreak{}column}---
and directed ChE relations---feed\_mixing feeds heater; heater feeds reactor; reactor feeds flash\_separator; flash\_separator vapor\_outlet\_feeds compressor; compressor recycles\_to feed\_mixing; flash\_separator liquid\_outlet\_feeds liquid\_preheater; liquid\_preheater feeds distillation\_column---to the downstream roles. Preserve explicit
negations and bounded assumptions; confidence 1.}

\tracestage{3}{Visual evidence: translate the PFD into an inspectable observation}

\tracetool{ChE evidence coordinator}{Visual Descriptor}{Provide a role-scoped engineering
evidence packet from the process request and authoritative process documentation. It contains
equipment functions, visible labels, material/energy-flow direction, recycle or branch cues,
operating semantics, and an explicit uncertainty boundary. LangGraph exposes this ChE evidence
to the perception role without adding executable implementation details.}

\traceagent{Visual Descriptor}{VisualGraphIR Validator / Topology Agent}{
Act as the modeling-assistance perception adapter. The validated visual
record has 14 nodes and 14 edges.
Complete node inventory: \code{M101}=Mixer; \code{H101}=Heater; \code{R101}=CSTR; \code{F101}=Flash; \code{S101}=Splitter; \code{C101}=PressureChanger; \code{Translator}=Translator; \code{H102}=Heater; \code{D101}=TrayColumn; \code{Feed\_\allowbreak{}H2\_\allowbreak{}CH4}=External Feed; \code{Feed\_\allowbreak{}Toluene}=External Feed; \code{Purge}=External Product; \code{Benzene}=External Product; \code{Toluene}=External Product. Complete directed-edge inventory: \code{s03}: \code{Feed\_\allowbreak{}H2\_\allowbreak{}CH4}$\rightarrow$\code{M101}; \code{s04}: \code{Feed\_\allowbreak{}Toluene}$\rightarrow$\code{M101}; \code{s05}: \code{M101}$\rightarrow$\code{H101}; \code{s06}: \code{H101}$\rightarrow$\code{R101}; \code{s07}: \code{R101}$\rightarrow$\code{F101}; \code{s08}: \code{F101}$\rightarrow$\code{S101}; \code{s09}: \code{C101}$\rightarrow$\code{M101} (recycle cue); \code{s10a}: \code{F101}$\rightarrow$\code{Translator}; \code{s10b}: \code{Translator}$\rightarrow$\code{H102}; \code{s11}: \code{H102}$\rightarrow$\code{D101}; \code{s12}: \code{S101}$\rightarrow$\code{Purge}; \code{s13}: \code{S101}$\rightarrow$\code{C101}; \code{s14}: \code{D101}$\rightarrow$\code{Benzene}; \code{s15}: \code{D101}$\rightarrow$\code{Toluene}.
Visible text labels: none.
Uncertainties: []. Observation excerpt:
\begin{tracejsonbox}
\{\\
  "schema\_version": "visual\_graph\_ir/2",\\
  "nodes": [\\
    \{\\
      "visual\_id": "M101",\\
      "label\_text": "M101",\\
      "equipment\_class\_guess": "Mixer",\\
      "bbox": [\\
        0.12,\\
        0.35,\\
        0.24,\\
        0.5\\
      ],\\
      "confidence\_0\_1": 0.95,\\
      "evidence\_ref": "triangle mixer symbol with three inlet arrows"\\
    \}\\
  ],\\
  "edges": [\\
    \{\\
      "visual\_edge\_id": "s03",\\
      "source\_visual\_id": "Feed\_H2\_CH4",\\
      "destination\_visual\_id": "M101",\\
      "label\_text": "",\\
      "direction\_confidence\_0\_1": 0.95,\\
      "is\_recycle\_guess": false,\\
      "evidence\_ref": "arrow from feed box to mixer"\\
    \}\\
  ],\\
  "uncertainties": []\\
\}\\
\end{tracejsonbox}
The Topology Agent combines this typed observation with routed ChE constraints and simulator
capabilities; visual perception alone does not choose the executable flowsheet.}

\tracetool{VisualGraphIR Validator}{Topology Agent}{Validate the visual schema, unique
node/edge identifiers, normalized geometry, endpoint closure, evidence references, and
uncertainty boundary before topology synthesis.}

\tracestage{4}{Topology: combine observations with ChE and simulator constraints}

\traceagent{Topology Agent}{Topology Validator / Prebuilder}{Combine the typed visual record
with routed ChE constraints and simulator capabilities. Emit all
9 typed units: \code{M101} Mixer (ChE role mix toluene feed, hydrogen feed, and vapor recycle, package \code{BTHM\_\allowbreak{}params}, stage \code{recycle\_\allowbreak{}section}); \code{H101} Heater (ChE role reactor preheater, package \code{BTHM\_\allowbreak{}params}, stage \code{recycle\_\allowbreak{}section}); \code{R101} CSTR (ChE role HDA reactor, package \code{BTHM\_\allowbreak{}params}, stage \code{recycle\_\allowbreak{}section}); \code{F101} Flash (ChE role separate non-condensibles from liquid benzene/toluene, package \code{BTHM\_\allowbreak{}params}, stage \code{recycle\_\allowbreak{}section}); \code{S101} Splitter (ChE role split F101 vapor into purge and recycle, package \code{BTHM\_\allowbreak{}params}, stage \code{recycle\_\allowbreak{}section}); \code{C101} PressureChanger (ChE role compress vapor recycle before M101, package \code{BTHM\_\allowbreak{}params}, stage \code{recycle\_\allowbreak{}section}); \code{translator} Translator (ChE role map HDA liquid to two-component BTX package, package \code{BTHM\_\allowbreak{}params\_\allowbreak{}to\_\allowbreak{}BT\_\allowbreak{}params}, stage \code{distillation\allowbreak{}\_\allowbreak{}section}); \code{H102} Heater (ChE role condition BTX liquid stream before distillation, package \code{BT\_\allowbreak{}params}, stage \code{distillation\allowbreak{}\_\allowbreak{}section}); \code{D101} TrayColumn (ChE role benzene/toluene distillation column, package \code{BT\_\allowbreak{}params}, stage \code{distillation\allowbreak{}\_\allowbreak{}section}). Property/reaction package
surface: \code{BTHM\_\allowbreak{}params} (components=[benzene, toluene, hydrogen, methane]; phases=[Liq, Vap]; scope=[M101, H101, R101, F101, S101, C101, translator.inlet]); \code{BT\_\allowbreak{}params} (components=[benzene, toluene]; phases=[Liq, Vap]; scope=[translator.outlet, H102, D101]). Emit all 9 exact stream endpoints:
\code{s03}: \code{M101.\allowbreak{}outlet} $\rightarrow$ \code{H101.\allowbreak{}inlet}; \code{s04}: \code{H101.\allowbreak{}outlet} $\rightarrow$ \code{R101.\allowbreak{}inlet}; \code{s05}: \code{R101.\allowbreak{}outlet} $\rightarrow$ \code{F101.\allowbreak{}inlet}; \code{s06}: \code{F101.\allowbreak{}vap\_\allowbreak{}outlet} $\rightarrow$ \code{S101.\allowbreak{}inlet}; \code{s08}: \code{S101.\allowbreak{}recycle} $\rightarrow$ \code{C101.\allowbreak{}inlet}; \code{s09}: \code{C101.\allowbreak{}outlet} $\rightarrow$ \code{M101.\allowbreak{}vapor\_\allowbreak{}recycle}; \code{s10}: \code{F101.\allowbreak{}liq\_\allowbreak{}outlet} $\rightarrow$ \code{translator.\allowbreak{}inlet}; \code{s11\_\allowbreak{}pre\_\allowbreak{}column}: \code{translator.\allowbreak{}outlet} $\rightarrow$ \code{H102.\allowbreak{}inlet}; \code{s11}: \code{H102.\allowbreak{}outlet} $\rightarrow$ \code{D101.\allowbreak{}feed}. Terminal feeds: \code{M101.\allowbreak{}toluene\_\allowbreak{}feed}; \code{M101.\allowbreak{}hydrogen\_\allowbreak{}feed};
terminal products: \code{S101.\allowbreak{}purge}; \code{D101.\allowbreak{}condenser.\allowbreak{}distillate}; \code{D101.\allowbreak{}reboiler.\allowbreak{}bottoms}.}

\tracestage{5}{Specification: close operating choices and degrees of freedom}

\tracetool{Topology Validator / Prebuilder}{Specification Agent}{The typed unit,
package, configuration, port, arc, terminal, and construction checks pass for the accepted
TopologyIR; only the accepted topology advances.}

\traceagent{Specification Agent}{DOF Checker}{Emit all 19 persisted SpecIR
assignments one-to-one. The provenance marker \emph{ChE-agent/DoF-validated} denotes a
Specification Agent proposal accepted by the deterministic DoF gate.
\textbf{Unit configuration (8)}:
\code{R101.\allowbreak{}conversion} = \code{0.\allowbreak{}75} dimensionless (toluene conversion target; ChE-agent/DoF-validated); \code{R101.\allowbreak{}heat\_\allowbreak{}duty[\allowbreak{}0]\allowbreak{}} = \code{0} W (adiabatic reactor duty; ChE-agent/DoF-validated); \code{F101.\allowbreak{}deltaP[\allowbreak{}0]\allowbreak{}} = \code{0} Pa (flash pressure drop; ChE-agent/DoF-validated); \code{S101.\allowbreak{}split\_\allowbreak{}fraction[\allowbreak{}0, 'purge']\allowbreak{}} = \code{0.\allowbreak{}2} dimensionless (purge split fraction; ChE-agent/DoF-validated); \code{H102.\allowbreak{}deltaP[\allowbreak{}0]\allowbreak{}} = \code{-\allowbreak{}200000} Pa (BTX heater pressure drop; ChE-agent/DoF-validated); \code{D101.\allowbreak{}condenser.\allowbreak{}reflux\_\allowbreak{}ratio} = \code{0.\allowbreak{}5} dimensionless (column reflux ratio; ChE-agent/DoF-validated); \code{D101.\allowbreak{}reboiler.\allowbreak{}boilup\_\allowbreak{}ratio} = \code{0.\allowbreak{}5} dimensionless (column boilup ratio; ChE-agent/DoF-validated); \code{D101.\allowbreak{}condenser.\allowbreak{}condenser\_\allowbreak{}pressure} = \code{150000} Pa (column condenser pressure; ChE-agent/DoF-validated). \textbf{Stream configuration (11)}:
\code{M101.\allowbreak{}toluene\_\allowbreak{}feed.\allowbreak{}flow\_\allowbreak{}mol\_\allowbreak{}phase\_\allowbreak{}comp[\allowbreak{}0, 'Liq', 'toluene']\allowbreak{}} = \code{0.\allowbreak{}3} mol/s (toluene feed molar flow; ChE-agent/DoF-validated); \code{M101.\allowbreak{}hydrogen\_\allowbreak{}feed.\allowbreak{}flow\_\allowbreak{}mol\_\allowbreak{}phase\_\allowbreak{}comp[\allowbreak{}0, 'Vap', 'hydrogen']\allowbreak{}} = \code{0.\allowbreak{}3} mol/s (hydrogen feed molar flow; ChE-agent/DoF-validated); \code{M101.\allowbreak{}hydrogen\_\allowbreak{}feed.\allowbreak{}flow\_\allowbreak{}mol\_\allowbreak{}phase\_\allowbreak{}comp[\allowbreak{}0, 'Vap', 'methane']\allowbreak{}} = \code{0.\allowbreak{}02} mol/s (methane impurity in hydrogen feed; ChE-agent/DoF-validated); \code{M101.\allowbreak{}toluene\_\allowbreak{}feed.\allowbreak{}temperature[\allowbreak{}0]\allowbreak{}} = \code{303.\allowbreak{}2} K (toluene feed temperature; ChE-agent/DoF-validated); \code{M101.\allowbreak{}hydrogen\_\allowbreak{}feed.\allowbreak{}temperature[\allowbreak{}0]\allowbreak{}} = \code{303.\allowbreak{}2} K (hydrogen feed temperature; ChE-agent/DoF-validated); \code{M101.\allowbreak{}toluene\_\allowbreak{}feed.\allowbreak{}pressure[\allowbreak{}0]\allowbreak{}} = \code{350000} Pa (toluene feed pressure; ChE-agent/DoF-validated); \code{M101.\allowbreak{}hydrogen\_\allowbreak{}feed.\allowbreak{}pressure[\allowbreak{}0]\allowbreak{}} = \code{350000} Pa (hydrogen feed pressure; ChE-agent/DoF-validated); \code{H101.\allowbreak{}outlet.\allowbreak{}temperature[\allowbreak{}0]\allowbreak{}} = \code{600} K (reactor preheat target; ChE-agent/DoF-validated); \code{F101.\allowbreak{}vap\_\allowbreak{}outlet.\allowbreak{}temperature[\allowbreak{}0]\allowbreak{}} = \code{325} K (first flash vapor temperature; ChE-agent/DoF-validated); \code{C101.\allowbreak{}outlet.\allowbreak{}pressure[\allowbreak{}0]\allowbreak{}} = \code{350000} Pa (recycle compressor outlet pressure; ChE-agent/DoF-validated); \code{H102.\allowbreak{}outlet.\allowbreak{}temperature[\allowbreak{}0]\allowbreak{}} = \code{375} K (BTX column feed conditioning target; ChE-agent/DoF-validated). No assignment is replaced by a count.}

\tracetool{DOF Checker}{BuildPlan Compiler}{Validate target existence, compatibility,
duplicate/conflicting fixes, terminal closure, and expected steady-state DoF. The promoted
SpecIR and deterministic DoF report pass before compilation.}

\tracestage{6}{Build and initialize the accepted white-box model}

\tracetool{BuildPlan Compiler}{Initialization / Native Solver}{Compile the hermetic
BuildPlan from the accepted TopologyIR and SpecIR. Initialization contract:
solve order=[M101, H101, R101, F101, S101, C101, translator, H102, D101]; strategy=SequentialDecomposition over recycle section, then add/initialize TrayColumn. Recycle/tear contract: []. Solve contract:
native\_solve\_disabled=false; expected\_dof=0.}

\tracestage{7}{Solve, inspect diagnostics, repair, and optimize only when eligible}

\tracetool{Native Solver}{Failure Router / Report}{Persist pass=\code{True},
termination \code{optimal}, final DoF
\code{0}, stage \code{steady\_\allowbreak{}state\_\allowbreak{}solver}, and no terminal error.
Representative same-run stream/product quantities: \code{s11} [flow\_mol=0.267119; temperature=375; pressure=150,000; mole\_frac\_comp: benzene=0.765945, toluene=0.234055]; \code{H102\_\allowbreak{}outlet\_\allowbreak{}to\_\allowbreak{}column\_\allowbreak{}feed} [flow\_mol=0.267119; temperature=375; pressure=150,000; mole\_frac\_comp: benzene=0.765945, toluene=0.234055]; \code{COL1\_\allowbreak{}condenser\_\allowbreak{}distillate} [flow\_mol=0.161969; temperature=368.929; pressure=150,000; mole\_frac\_comp: benzene=0.894916, toluene=0.105084]; \code{COL1\_\allowbreak{}reboiler\_\allowbreak{}bottoms} [flow\_mol=0.10515; temperature=377.296; pressure=150,000; mole\_frac\_comp: benzene=0.566774, toluene=0.433226]. Source
evidence is reported as the accepted output of the native execution node.}

\traceagent{Debug Supervisor audit (standby)}{Failure Router / Report}{No fault was routed to
Debug; no diagnosis, repair artifact, or replay was generated. The supervisor remained armed
from before Stage 1 through terminal acceptance.}

\tracetool{Failure Router}{Optimization Agent / Report}{Read the structured topology,
specification, DoF, and accepted execution reports. Close the bounded repair branch, then pass the
accepted simulator state to Optimization only when the request makes that role eligible.}

\traceagent{Optimization Agent}{Deterministic runner / Report}{Optimization was not invoked because the normalized intent \code{build\_\allowbreak{}and\_\allowbreak{}solve} does not request a separate design-optimization stage.}

\tracetool{IDAES FV exporter}{Web / manuscript report}{After solve and any eligible
optimization, export the same-run flowsheet visualization with 26 FV
cells. Retain the ChE process evidence, accepted topology, FV graph, and terminal numerical
evidence as one LangGraph trajectory.}

\end{tracecasebox}

\casefigure{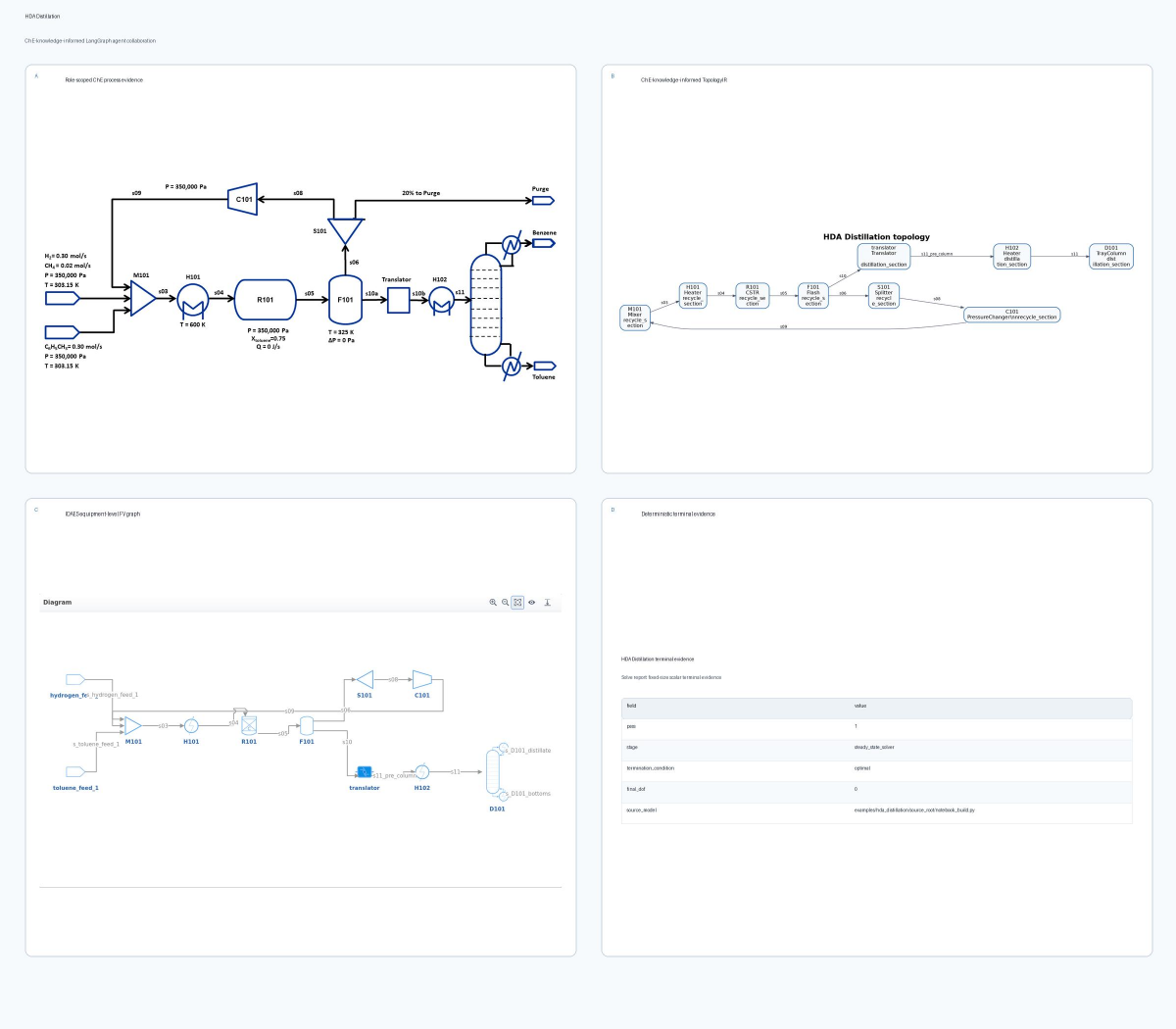}
{HDA distillation collaboration audit bundle.}
{fig:arxiv-case-hda-distillation}

\subsubsection{Methanol Recycle}
\begin{tracecasebox}{Methanol recycle---complete interaction trace}
\scriptsize
\runbar{}

\tracetool{LangGraph collaboration profile}{Case report}{Input Understanding, Intent,
Visual, Topology, Specification, deterministic validation, execution, and conditional
Debug/Optimization roles form one artifact-consistent trajectory. Role reasoning is bounded by
medium. The visibility gateway records 5 scoped model-agent
calls and 15 visibility-audit events with zero boundary errors.
Structured-output repairs=0; Debug replays=0. The
Debug supervisor is armed before Stage 1. The deterministic execution returns \code{solved} with
final acceptance at the report boundary.}

\tracestage{1}{Input Understanding: preserve requirements, relations, and negations}

\traceuser{Root User}{Input Understanding}{Build a methanol-synthesis flowsheet from hydrogen and carbon monoxide with feed mixing, compression, heating, reaction, cooling, flash separation, purge, and vapor recycle to the reactor feed.}

\traceagent{Input Understanding}{Intent Router}{Normalize the request as
\code{build\_\allowbreak{}and\_\allowbreak{}solve} in methanol\_synthesis. Required
features: \code{feed\_\allowbreak{}mixing}; \code{compressor}; \code{heater}; \code{reactor}; \code{cooler}; \code{flash\_\allowbreak{}separator}; \code{purge\_\allowbreak{}stream}; \code{vapor\_\allowbreak{}recycle\_\allowbreak{}loop}. Negated features:
none. Directed relations: fresh\_feed feeds feed\_mixer; feed\_mixer feeds compressor; compressor feeds heater; heater feeds reactor; reactor feeds cooler; cooler feeds flash\_separator; flash\_vapor\_outlet splits\_to purge\_stream; flash\_vapor\_outlet splits\_to recycle\_compressor; recycle\_compressor feeds feed\_mixer.
Assumptions: Standard methanol synthesis topology assumed: fresh feed mixes with recycle before compression.;
ambiguities: none;
clarification required=\code{False}; confidence
0.95.}

\tracestage{2}{Intent: recover the ChE process boundary and downstream obligations}

\traceagent{Intent Router}{ChE evidence coordinator / Visual Descriptor}{Interpret the
request as Methanol Recycle in the reaction domain.
Route the required equipment and process features---\code{feed\_\allowbreak{}mixing}; \code{compressor}; \code{heater}; \code{reactor}; \code{cooler}; \code{flash\_\allowbreak{}separator}; \code{purge\_\allowbreak{}stream}; \code{vapor\_\allowbreak{}recycle\_\allowbreak{}loop}---
and directed ChE relations---fresh\_feed feeds feed\_mixer; feed\_mixer feeds compressor; compressor feeds heater; heater feeds reactor; reactor feeds cooler; cooler feeds flash\_separator; flash\_vapor\_outlet splits\_to purge\_stream; flash\_vapor\_outlet splits\_to recycle\_compressor; recycle\_compressor feeds feed\_mixer---to the downstream roles. Preserve explicit
negations and bounded assumptions; confidence 1.}

\tracestage{3}{Visual evidence: translate the PFD into an inspectable observation}

\tracetool{ChE evidence coordinator}{Visual Descriptor}{Provide a role-scoped engineering
evidence packet from the process request and authoritative process documentation. It contains
equipment functions, visible labels, material/energy-flow direction, recycle or branch cues,
operating semantics, and an explicit uncertainty boundary. LangGraph exposes this ChE evidence
to the perception role without adding executable implementation details.}

\traceagent{Visual Descriptor}{VisualGraphIR Validator / Topology Agent}{
Act as the modeling-assistance perception adapter. The validated visual
record has 13 nodes and 13 edges.
Complete node inventory: \code{H2\_\allowbreak{}feed}=Feed; \code{CO\_\allowbreak{}feed}=Feed; \code{M101}=Mixer; \code{M102}=Mixer; \code{C101}=Compressor; \code{H101}=Heater; \code{R101}=StoichiometricReactor; \code{T101}=Turbine; \code{H102}=Heater; \code{F101}=Flash; \code{S101}=Splitter; \code{EXHAUST}=Product; \code{CH3OH\_\allowbreak{}product}=Product. Complete directed-edge inventory: \code{s01}: \code{H2\_\allowbreak{}feed}$\rightarrow$\code{M101}; \code{s02}: \code{CO\_\allowbreak{}feed}$\rightarrow$\code{M101}; \code{s03}: \code{M101}$\rightarrow$\code{M102}; \code{s04}: \code{M102}$\rightarrow$\code{C101}; \code{s05}: \code{C101}$\rightarrow$\code{H101}; \code{s06}: \code{H101}$\rightarrow$\code{R101}; \code{s07}: \code{R101}$\rightarrow$\code{T101}; \code{s08}: \code{T101}$\rightarrow$\code{H102}; \code{s09}: \code{H102}$\rightarrow$\code{F101}; \code{s10}: \code{F101}$\rightarrow$\code{S101}; \code{s11}: \code{S101}$\rightarrow$\code{EXHAUST}; \code{s12}: \code{S101}$\rightarrow$\code{M102} (recycle cue); \code{s13}: \code{F101}$\rightarrow$\code{CH3OH\_\allowbreak{}product}.
Visible text labels: none.
Uncertainties: []. Observation excerpt:
\begin{tracejsonbox}
\{\\
  "schema\_version": "visual\_graph\_ir/2",\\
  "nodes": [\\
    \{\\
      "visual\_id": "H2\_feed",\\
      "label\_text": "H2, 20 C, 30 bar",\\
      "equipment\_class\_guess": "Feed",\\
      "bbox": [\\
        0.0,\\
        0.15,\\
        0.12,\\
        0.3\\
      ],\\
      "confidence\_0\_1": 0.95,\\
      "evidence\_ref": "Leftmost labeled arrow with H2 specification"\\
    \}\\
  ],\\
  "edges": [\\
    \{\\
      "visual\_edge\_id": "s01",\\
      "source\_visual\_id": "H2\_feed",\\
      "destination\_visual\_id": "M101",\\
      "label\_text": "",\\
      "direction\_confidence\_0\_1": 0.95,\\
      "is\_recycle\_guess": false,\\
      "evidence\_ref": "Arrow from H2\_feed to M101"\\
    \}\\
  ],\\
  "uncertainties": []\\
\}\\
\end{tracejsonbox}
The Topology Agent combines this typed observation with routed ChE constraints and simulator
capabilities; visual perception alone does not choose the executable flowsheet.}

\tracetool{VisualGraphIR Validator}{Topology Agent}{Validate the visual schema, unique
node/edge identifiers, normalized geometry, endpoint closure, evidence references, and
uncertainty boundary before topology synthesis.}

\tracestage{4}{Topology: combine observations with ChE and simulator constraints}

\traceagent{Topology Agent}{Topology Validator / Prebuilder}{Combine the typed visual record
with routed ChE constraints and simulator capabilities. Emit all
13 typed units: \code{H2} Feed (ChE role hydrogen feed, package \code{thermo\_\allowbreak{}params\_\allowbreak{}vapor}, stage \code{feed\_\allowbreak{}section}); \code{CO} Feed (ChE role carbon monoxide feed, package \code{thermo\_\allowbreak{}params\_\allowbreak{}vapor}, stage \code{feed\_\allowbreak{}section}); \code{M101} Mixer (ChE role mix H2 and CO feeds, package \code{thermo\_\allowbreak{}params\_\allowbreak{}vapor}, stage \code{feed\_\allowbreak{}section}); \code{M102} Mixer (ChE role mix fresh syngas with recycle vapor, package \code{thermo\_\allowbreak{}params\_\allowbreak{}vapor}, stage \code{recycle\_\allowbreak{}section}); \code{C101} Compressor (ChE role compress mixed syngas, package \code{thermo\_\allowbreak{}params\_\allowbreak{}vapor}, stage \code{reaction\_\allowbreak{}section}); \code{H101} Heater (ChE role preheat reactor feed, package \code{thermo\_\allowbreak{}params\_\allowbreak{}vapor}, stage \code{reaction\_\allowbreak{}section}); \code{R101} StoichiometricReactor (ChE role methanol synthesis reactor, package \code{thermo\_\allowbreak{}params\_\allowbreak{}vapor}, stage \code{reaction\_\allowbreak{}section}); \code{T101} Turbine (ChE role expand reactor effluent, package \code{thermo\_\allowbreak{}params\_\allowbreak{}vapor}, stage \code{recovery\_\allowbreak{}section}); \code{H102} Heater (ChE role cool product recovery feed, package \code{thermo\_\allowbreak{}params\_\allowbreak{}vapor}, stage \code{recovery\_\allowbreak{}section}); \code{F101} Flash (ChE role recover liquid methanol from vapor, package \code{thermo\_\allowbreak{}params\_\allowbreak{}VLE}, stage \code{recovery\_\allowbreak{}section}); \code{S101} Splitter (ChE role split flash vapor into purge and recycle, package \code{thermo\_\allowbreak{}params\_\allowbreak{}vapor}, stage \code{recycle\_\allowbreak{}section}); \code{EXHAUST} Product (ChE role purge gas product, package \code{thermo\_\allowbreak{}params\_\allowbreak{}vapor}, stage \code{product\_\allowbreak{}section}); \code{CH3OH} Product (ChE role liquid methanol product, package \code{thermo\_\allowbreak{}params\_\allowbreak{}VLE}, stage \code{product\_\allowbreak{}section}). Property/reaction package
surface: \code{thermo\_\allowbreak{}params\_\allowbreak{}VLE} (components=[H2, CO, CH3OH, CH4]; phases=[Liq, Vap]; scope=[F101, CH3OH]); \code{thermo\_\allowbreak{}params\_\allowbreak{}vapor} (components=[H2, CO, CH3OH, CH4]; phases=[Vap]; scope=[H2, CO, M101, M102, C101, H101, R101, T101, H102, S101, EXHAUST]). Emit all 13 exact stream endpoints:
\code{s01}: \code{H2.\allowbreak{}outlet} $\rightarrow$ \code{M101.\allowbreak{}H2\_\allowbreak{}WGS}; \code{s02}: \code{CO.\allowbreak{}outlet} $\rightarrow$ \code{M101.\allowbreak{}CO\_\allowbreak{}WGS}; \code{s03}: \code{M101.\allowbreak{}outlet} $\rightarrow$ \code{M102.\allowbreak{}feed}; \code{s04}: \code{M102.\allowbreak{}outlet} $\rightarrow$ \code{C101.\allowbreak{}inlet}; \code{s05}: \code{C101.\allowbreak{}outlet} $\rightarrow$ \code{H101.\allowbreak{}inlet}; \code{s06}: \code{H101.\allowbreak{}outlet} $\rightarrow$ \code{R101.\allowbreak{}inlet}; \code{s07}: \code{R101.\allowbreak{}outlet} $\rightarrow$ \code{T101.\allowbreak{}inlet}; \code{s08}: \code{T101.\allowbreak{}outlet} $\rightarrow$ \code{H102.\allowbreak{}inlet}; \code{s09}: \code{H102.\allowbreak{}outlet} $\rightarrow$ \code{F101.\allowbreak{}inlet}; \code{s10}: \code{F101.\allowbreak{}vap\_\allowbreak{}outlet} $\rightarrow$ \code{S101.\allowbreak{}inlet}; \code{s11}: \code{S101.\allowbreak{}recycle} $\rightarrow$ \code{M102.\allowbreak{}recycle}; \code{s12}: \code{S101.\allowbreak{}purge} $\rightarrow$ \code{EXHAUST.\allowbreak{}inlet}; \code{s13}: \code{F101.\allowbreak{}liq\_\allowbreak{}outlet} $\rightarrow$ \code{CH3OH.\allowbreak{}inlet}. Terminal feeds: none;
terminal products: none.}

\tracestage{5}{Specification: close operating choices and degrees of freedom}

\tracetool{Topology Validator / Prebuilder}{Specification Agent}{The typed unit,
package, configuration, port, arc, terminal, and construction checks pass for the accepted
TopologyIR; only the accepted topology advances.}

\traceagent{Specification Agent}{DOF Checker}{Emit all 14 persisted SpecIR
assignments one-to-one. The provenance marker \emph{ChE-agent/DoF-validated} denotes a
Specification Agent proposal accepted by the deterministic DoF gate.
\textbf{Unit configuration (6)}:
\code{R101.\allowbreak{}co\_\allowbreak{}conversion} = \code{0.\allowbreak{}75} dimensionless (CO conversion; ChE-agent/DoF-validated); \code{T101.\allowbreak{}deltaP[\allowbreak{}0]\allowbreak{}} = \code{-\allowbreak{}2000000} Pa (turbine pressure drop; ChE-agent/DoF-validated); \code{T101.\allowbreak{}efficiency\_\allowbreak{}isentropic[\allowbreak{}0]\allowbreak{}} = \code{0.\allowbreak{}9} dimensionless (turbine isentropic efficiency; ChE-agent/DoF-validated); \code{F101.\allowbreak{}deltaP[\allowbreak{}0]\allowbreak{}} = \code{0} Pa (flash pressure drop; ChE-agent/DoF-validated); \code{F101.\allowbreak{}outlet.\allowbreak{}temperature[\allowbreak{}0]\allowbreak{}} = \code{407.\allowbreak{}15} K (flash outlet temperature; ChE-agent/DoF-validated); \code{S101.\allowbreak{}split\_\allowbreak{}fraction[\allowbreak{}0, 'purge']\allowbreak{}} = \code{0.\allowbreak{}9999} dimensionless (purge split fraction; ChE-agent/DoF-validated). \textbf{Stream configuration (8)}:
\code{H2.\allowbreak{}outlet.\allowbreak{}flow\_\allowbreak{}mol[\allowbreak{}0]\allowbreak{}} = \code{637.\allowbreak{}2} mol/s (hydrogen feed flow; ChE-agent/DoF-validated); \code{CO.\allowbreak{}outlet.\allowbreak{}flow\_\allowbreak{}mol[\allowbreak{}0]\allowbreak{}} = \code{316.\allowbreak{}8} mol/s (carbon monoxide feed flow; ChE-agent/DoF-validated); \code{H2.\allowbreak{}outlet.\allowbreak{}pressure[\allowbreak{}0]\allowbreak{}} = \code{3000000} Pa (hydrogen feed pressure; ChE-agent/DoF-validated); \code{CO.\allowbreak{}outlet.\allowbreak{}pressure[\allowbreak{}0]\allowbreak{}} = \code{3000000} Pa (carbon monoxide feed pressure; ChE-agent/DoF-validated); \code{C101.\allowbreak{}outlet.\allowbreak{}pressure[\allowbreak{}0]\allowbreak{}} = \code{5100000} Pa (compressor outlet pressure; ChE-agent/DoF-validated); \code{H101.\allowbreak{}outlet.\allowbreak{}temperature[\allowbreak{}0]\allowbreak{}} = \code{488.\allowbreak{}15} K (reactor feed preheat temperature; ChE-agent/DoF-validated); \code{R101.\allowbreak{}outlet.\allowbreak{}temperature[\allowbreak{}0]\allowbreak{}} = \code{507.\allowbreak{}15} K (reactor outlet temperature; ChE-agent/DoF-validated); \code{H102.\allowbreak{}outlet.\allowbreak{}temperature[\allowbreak{}0]\allowbreak{}} = \code{407.\allowbreak{}15} K (flash feed cooling temperature; ChE-agent/DoF-validated). No assignment is replaced by a count.}

\tracetool{DOF Checker}{BuildPlan Compiler}{Validate target existence, compatibility,
duplicate/conflicting fixes, terminal closure, and expected steady-state DoF. The promoted
SpecIR and deterministic DoF report pass before compilation.}

\tracestage{6}{Build and initialize the accepted white-box model}

\tracetool{BuildPlan Compiler}{Initialization / Native Solver}{Compile the hermetic
BuildPlan from the accepted TopologyIR and SpecIR. Initialization contract:
solve order=[H2, CO, M101, M102, C101, H101, R101, T101, H102, F101, S101]; strategy=SequentialDecomposition over methanol recycle loop. Recycle/tear contract: []. Solve contract:
native\_solve\_disabled=false; expected\_dof=0.}

\tracestage{7}{Solve, inspect diagnostics, repair, and optimize only when eligible}

\tracetool{Native Solver}{Failure Router / Report}{Persist pass=\code{True},
termination \code{optimal}, final DoF
\code{0}, stage \code{steady\_\allowbreak{}state\_\allowbreak{}solver}, and no terminal error.
Representative same-run stream/product quantities: \code{s10\_\allowbreak{}flash\_\allowbreak{}vapor} [flow\_mol=336.239; temperature=not recorded; pressure=3.1e+06; enth\_mol=-80217.7; mole\_frac\_comp: H2=0.481811, CO=0.235553, CH3OH=0.282634, CH4=2.83755e-06]; \code{s11\_\allowbreak{}recycle} [flow\_mol=0.0336239; temperature=not recorded; pressure=3.1e+06; enth\_mol=-80217.7; mole\_frac\_comp: H2=0.481811, CO=0.235553, CH3OH=0.282634, CH4=2.83755e-06]; \code{s12\_\allowbreak{}purge} [flow\_mol=336.206; temperature=not recorded; pressure=3.1e+06; enth\_mol=-80217.7; mole\_frac\_comp: H2=0.481811, CO=0.235553, CH3OH=0.282634, CH4=2.83755e-06]; \code{s13\_\allowbreak{}methanol\_\allowbreak{}product} [flow\_mol=142.584; temperature=not recorded; pressure=3.1e+06; enth\_mol=-238,130; mole\_frac\_comp: H2=1e-08, CO=1e-08, CH3OH=1, CH4=1e-08]. Source
evidence is reported as the accepted output of the native execution node.}

\traceagent{Debug Supervisor audit (standby)}{Failure Router / Report}{No fault was routed to
Debug; no diagnosis, repair artifact, or replay was generated. The supervisor remained armed
from before Stage 1 through terminal acceptance.}

\tracetool{Failure Router}{Optimization Agent / Report}{Read the structured topology,
specification, DoF, and accepted execution reports. Close the bounded repair branch, then pass the
accepted simulator state to Optimization only when the request makes that role eligible.}

\traceagent{Optimization Agent}{Deterministic runner / Report}{Optimization was not invoked because the normalized intent \code{build\_\allowbreak{}and\_\allowbreak{}solve} does not request a separate design-optimization stage.}

\tracetool{IDAES FV exporter}{Web / manuscript report}{After solve and any eligible
optimization, export the same-run flowsheet visualization with 26 FV
cells. Retain the ChE process evidence, accepted topology, FV graph, and terminal numerical
evidence as one LangGraph trajectory.}

\end{tracecasebox}

\casefigure{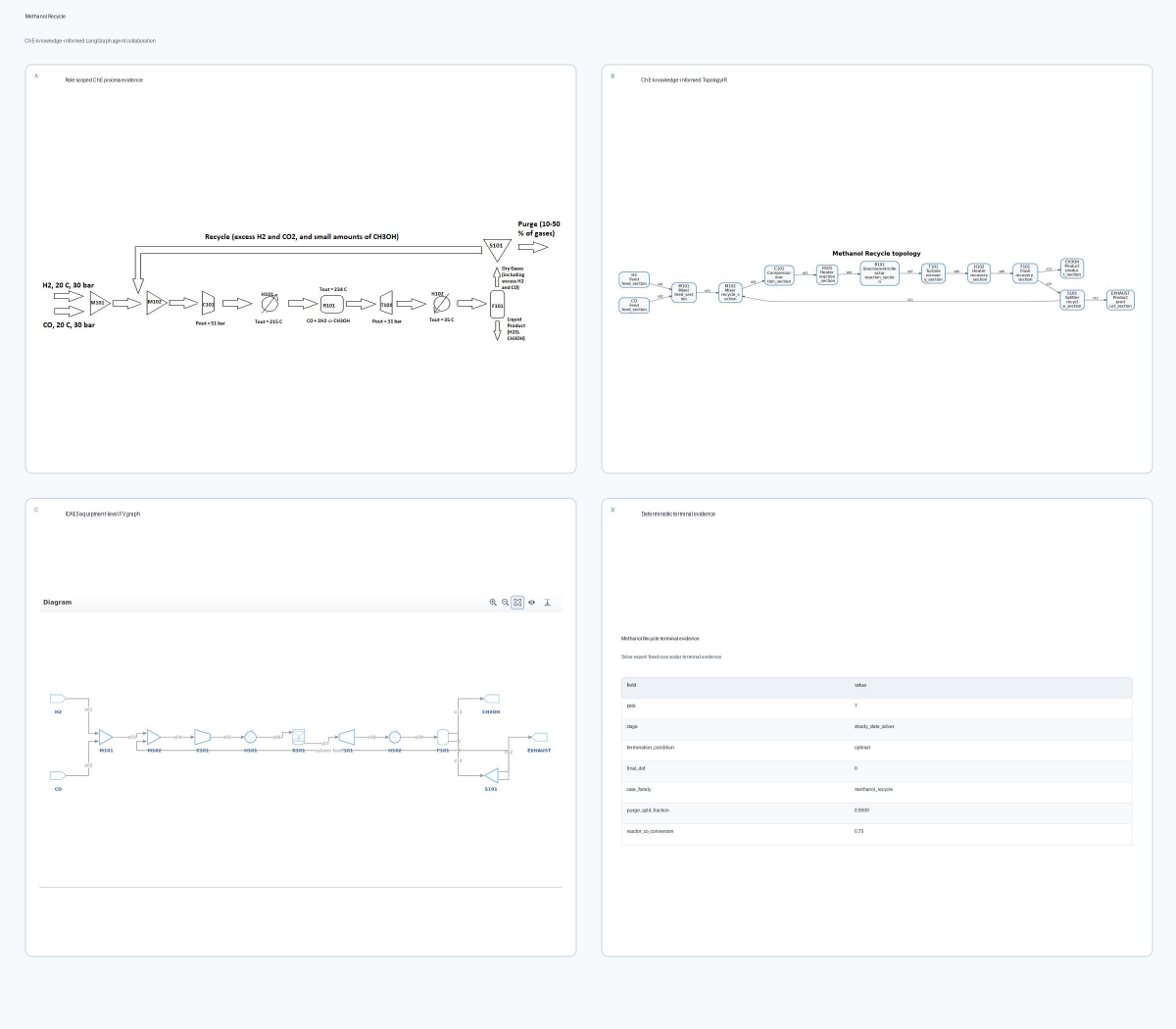}
{Methanol recycle collaboration audit bundle.}
{fig:arxiv-case-methanol-recycle}

\subsubsection{Membrane Distillation Recirculation}
\begin{tracecasebox}{Membrane distillation recirculation: complete interaction trace}
\scriptsize
\runbar{}

\tracetool{LangGraph collaboration profile}{Case report}{Input Understanding, Intent,
Visual, Topology, Specification, deterministic validation, execution, and conditional
Debug/Optimization roles form one artifact-consistent trajectory. Role reasoning is bounded by
medium. The visibility gateway records 6 scoped model-agent
calls and 18 visibility-audit events with zero boundary errors.
Structured-output repairs=0; Debug replays=0. The
Debug supervisor is armed before Stage 1. The deterministic execution returns \code{optimized} with
final acceptance at the report boundary.}

\tracestage{1}{Input Understanding: preserve requirements, relations, and negations}

\traceuser{Root User}{Input Understanding}{Build a WaterTAP direct-contact membrane-distillation flowsheet with feed pumping, heat recovery, brine heating, the MD unit, concentrate recycle and reject handling, permeate cooling, and a final permeate product. After the zero-DoF steady-state solve, perform the registered deterministic design optimization and retain its objective, decision variables, bounds, constraints, termination, and terminal metrics in this same run.}

\traceagent{Input Understanding}{Intent Router}{Normalize the request as
\code{optimize} in WaterTAP membrane distillation. Required
features: \code{feed\_\allowbreak{}pump}; \code{heat\_\allowbreak{}recovery\_\allowbreak{}exchanger}; \code{brine\_\allowbreak{}heater}; \code{membrane\_\allowbreak{}distillation\allowbreak{}\_\allowbreak{}unit}; \code{concentrate\_\allowbreak{}recycle}; \code{reject\_\allowbreak{}handling}; \code{permeate\_\allowbreak{}cooler}; \code{permeate\_\allowbreak{}product}; \code{deterministi\allowbreak{}c\_\allowbreak{}design\_\allowbreak{}optimization\allowbreak{}}. Negated features:
none. Directed relations: feed\_inlet feeds feed\_pump; feed\_pump feeds heat\_recovery\_exchanger\_feed\_side; heat\_recovery\_exchanger\_feed\_side feeds brine\_heater; brine\_heater feeds membrane\_distillation\_unit\_feed; membrane\_distillation\_unit\_concentrate splits\_to concentrate\_recycle\_stream; membrane\_distillation\_unit\_concentrate splits\_to reject\_stream; concentrate\_recycle\_stream recycles\_to feed\_mixing\_point; membrane\_distillation\_unit\_permeate feeds permeate\_cooler; permeate\_cooler feeds permeate\_product.
Assumptions: Standard WaterTAP DCMD topology assumed for component ordering.; Optimization is performed on the same flowsheet instance after initial solve.;
ambiguities: none;
clarification required=\code{False}; confidence
0.95.}

\tracestage{2}{Intent: recover the ChE process boundary and downstream obligations}

\traceagent{Intent Router}{ChE evidence coordinator / Visual Descriptor}{Interpret the
request as Membrane Distillation in the distillation domain.
Route the required equipment and process features, namely \code{feed\_\allowbreak{}pump}; \code{heat\_\allowbreak{}recovery\_\allowbreak{}exchanger}; \code{brine\_\allowbreak{}heater}; \code{membrane\_\allowbreak{}distillation\allowbreak{}\_\allowbreak{}unit}; \code{concentrate\_\allowbreak{}recycle}; \code{reject\_\allowbreak{}handling}; \code{permeate\_\allowbreak{}cooler}; \code{permeate\_\allowbreak{}product}; \code{deterministi\allowbreak{}c\_\allowbreak{}design\_\allowbreak{}optimization\allowbreak{}},
and directed ChE relations, namely feed\_inlet feeds feed\_pump; feed\_pump feeds heat\_recovery\_exchanger\_feed\_side; heat\_recovery\_exchanger\_feed\_side feeds brine\_heater; brine\_heater feeds membrane\_distillation\_unit\_feed; membrane\_distillation\_unit\_concentrate splits\_to concentrate\_recycle\_stream; membrane\_distillation\_unit\_concentrate splits\_to reject\_stream; concentrate\_recycle\_stream recycles\_to feed\_mixing\_point; membrane\_distillation\_unit\_permeate feeds permeate\_cooler; permeate\_cooler feeds permeate\_product, to the downstream roles. Preserve explicit
negations and bounded assumptions; confidence 0.98.}

\tracestage{3}{Visual evidence: translate the PFD into an inspectable observation}

\tracetool{ChE evidence coordinator}{Visual Descriptor}{Provide a role-scoped engineering
evidence packet from the process request and authoritative process documentation. It contains
equipment functions, visible labels, material/energy-flow direction, recycle or branch cues,
operating semantics, and an explicit uncertainty boundary. LangGraph exposes this ChE evidence
to the perception role without adding executable implementation details.}

\traceagent{Visual Descriptor}{VisualGraphIR Validator / Topology Agent}{
Act as the modeling-assistance perception adapter. The validated visual
record has 8 nodes and 9 edges.
Complete node inventory: \code{Heater}=Heater; \code{Heat\_\allowbreak{}Exchanger}=HeatExchanger; \code{Cooler}=Heater; \code{MD\_\allowbreak{}Module}=MembraneDistillation0D; \code{Make\_\allowbreak{}up\_\allowbreak{}Feed}=Feed; \code{Reject\_\allowbreak{}Concentrate}=Product; \code{Permeate}=Product; \code{Recycle\_\allowbreak{}Stream}=Separator. Complete directed-edge inventory: \code{Make\_\allowbreak{}up\_\allowbreak{}Feed\_\allowbreak{}to\_\allowbreak{}MD\_\allowbreak{}hot\_\allowbreak{}inlet}: \code{Make\_\allowbreak{}up\_\allowbreak{}Feed}$\rightarrow$\code{MD\_\allowbreak{}Module}; \code{MD\_\allowbreak{}hot\_\allowbreak{}outlet\_\allowbreak{}to\_\allowbreak{}Reject\_\allowbreak{}Concentrate}: \code{MD\_\allowbreak{}Module}$\rightarrow$\code{Reject\_\allowbreak{}Concentrate}; \code{MD\_\allowbreak{}cold\_\allowbreak{}outlet\_\allowbreak{}to\_\allowbreak{}Heat\_\allowbreak{}Exchanger\_\allowbreak{}hot\_\allowbreak{}side}: \code{MD\_\allowbreak{}Module}$\rightarrow$\code{Heat\_\allowbreak{}Exchanger}; \code{Heat\_\allowbreak{}Exchanger\_\allowbreak{}cold\_\allowbreak{}side\_\allowbreak{}to\_\allowbreak{}Cooler}: \code{Heat\_\allowbreak{}Exchanger}$\rightarrow$\code{Cooler}; \code{Cooler\_\allowbreak{}to\_\allowbreak{}MD\_\allowbreak{}cold\_\allowbreak{}inlet}: \code{Cooler}$\rightarrow$\code{MD\_\allowbreak{}Module}; \code{Heat\_\allowbreak{}Exchanger\_\allowbreak{}hot\_\allowbreak{}side\_\allowbreak{}to\_\allowbreak{}Heater}: \code{Heat\_\allowbreak{}Exchanger}$\rightarrow$\code{Heater}; \code{Heater\_\allowbreak{}to\_\allowbreak{}MD\_\allowbreak{}hot\_\allowbreak{}inlet}: \code{Heater}$\rightarrow$\code{MD\_\allowbreak{}Module}; \code{MD\_\allowbreak{}cold\_\allowbreak{}outlet\_\allowbreak{}to\_\allowbreak{}Permeate\_\allowbreak{}Separator}: \code{MD\_\allowbreak{}Module}$\rightarrow$\code{Permeate}; \code{Recycle\_\allowbreak{}Stream\_\allowbreak{}from\_\allowbreak{}Reject\_\allowbreak{}Concentrate\_\allowbreak{}to\_\allowbreak{}Make\_\allowbreak{}up\_\allowbreak{}Feed}: \code{Reject\_\allowbreak{}Concentrate}$\rightarrow$\code{Make\_\allowbreak{}up\_\allowbreak{}Feed} (recycle cue).
Visible text labels: Make-up Feed; Reject Concentrate; Permeate; Recycle Stream; Coolant; Heater; Heat Exchanger; Cooler; Tf; Tfm; Tp; Cf; Cfm; qf; qm; qp; J.
Uncertainties: [\{'description': 'Internal stream splits within MD module are not visible; permeate and concentrate paths inferred from external outputs.', 'evidence\_ref': 'No explicit separators shown for permeate/concentrate split within MD module.'\}, \{'description': 'Exact connection point between Cooler output and MD cold inlet is ambiguous due to overlapping arrows.', 'evidence\_ref': 'Blue arrows converge near MD cold inlet but lack clear junction point.'\}]. Observation excerpt:
\begin{tracejsonbox}
\{\\
  "schema\_version": "visual\_graph\_ir/2",\\
  "nodes": [\\
    \{\\
      "visual\_id": "Heater",\\
      "label\_text": "Heater",\\
      "equipment\_class\_guess": "Heater",\\
      "bbox": [\\
        0.38,\\
        0.12,\\
        0.54,\\
        0.26\\
      ],\\
      "confidence\_0\_1": 0.95,\\
      "evidence\_ref": "Top circular unit labeled 'Heater'"\\
    \}\\
  ],\\
  "edges": [\\
    \{\\
      "visual\_edge\_id": "Make\_up\_Feed\_to\_MD\_hot\_inlet",\\
      "source\_visual\_id": "Make\_up\_Feed",\\
      "destination\_visual\_id": "MD\_Module",\\
      "label\_text": "",\\
      "direction\_confidence\_0\_1": 0.98,\\
      "is\_recycle\_guess": false,\\
      "evidence\_ref": "Red arrow from 'Make-up Feed' to MD hot inlet"\\
    \}\\
  ],\\
  "uncertainties": [\\
    "\{'description': 'Internal stream splits within MD module are not visible; permeate and concentrate paths inferred from external outputs.', 'evidence\_ref': 'No explicit separators shown for permeate/concentrate split within MD module.'\}",\\
    "\{'description': 'Exact connection point between Cooler output and MD cold inlet is ambiguous due to overlapping arrows.', 'evidence\_ref': 'Blue arrows converge near MD cold inlet but lack clear junction point.'\}"\\
  ]\\
\}\\
\end{tracejsonbox}
The Topology Agent combines this typed observation with routed ChE constraints and simulator
capabilities; visual perception alone does not choose the executable flowsheet.}

\tracetool{VisualGraphIR Validator}{Topology Agent}{Validate the visual schema, unique
node/edge identifiers, normalized geometry, endpoint closure, evidence references, and
uncertainty boundary before topology synthesis.}

\tracestage{4}{Topology: combine observations with ChE and simulator constraints}

\traceagent{Topology Agent}{Topology Validator / Prebuilder}{Combine the typed visual record
with routed ChE constraints and simulator capabilities. Emit all
13 typed units: \code{feed} Feed (ChE role seawater makeup feed, package \code{seawater}, stage \code{feed}); \code{pump\_\allowbreak{}feed} Pump (ChE role raise makeup feed pressure before mixing, package \code{seawater}, stage \code{hot\_\allowbreak{}loop}); \code{mixer} Mixer (ChE role mix makeup feed with concentrated brine recycle, package \code{seawater}, stage \code{hot\_\allowbreak{}loop}); \code{hx} HeatExchanger (ChE role recover heat from cold-loop outlet into hot-loop feed, package \code{seawater}, stage \code{heat\_\allowbreak{}recovery}); \code{pump\_\allowbreak{}brine} Pump (ChE role pressurize brine loop before heater, package \code{seawater}, stage \code{hot\_\allowbreak{}loop}); \code{heater} Heater (ChE role heat hot-side MD inlet stream, package \code{seawater}, stage \code{thermal}); \code{MD} MembraneDistillation0D (ChE role direct-contact membrane distillation module, package \code{seawater}, stage \code{separation}); \code{separator\_\allowbreak{}concentrate} Separator (ChE role split MD hot outlet into reject and recycle, package \code{seawater}, stage \code{hot\_\allowbreak{}loop}); \code{reject} Product (ChE role concentrated brine reject, package \code{seawater}, stage \code{product}); \code{chiller} Heater (ChE role cool cold-loop stream before MD cold inlet, package \code{permeate\_\allowbreak{}water}, stage \code{cold\_\allowbreak{}loop}); \code{separator\_\allowbreak{}permeate} Separator (ChE role split cold-loop outlet into product permeate and recycle loop, package \code{permeate\_\allowbreak{}water}, stage \code{cold\_\allowbreak{}loop}); \code{pump\_\allowbreak{}permeate} Pump (ChE role circulate cold-loop stream through chiller, package \code{permeate\_\allowbreak{}water}, stage \code{cold\_\allowbreak{}loop}); \code{permeate} Product (ChE role distilled permeate product, package \code{permeate\_\allowbreak{}water}, stage \code{product}). Property/reaction package
surface: \code{permeate\_\allowbreak{}water} (components=[H2O, TDS]; phases=[Liq]; scope=[chiller, MD.cold\_ch, hx.hot, separator\_permeate, pump\_permeate, permeate]); \code{seawater} (components=[H2O, TDS]; phases=[Liq]; scope=[feed, pump\_feed, mixer, hx.cold, pump\_brine, heater, MD.hot\_ch, separator\_concentrate, reject]); \code{water\_\allowbreak{}vapor} (components=[H2O]; phases=[Vap]; scope=[]). Emit all 15 exact stream endpoints:
\code{feed\_\allowbreak{}to\_\allowbreak{}feed\_\allowbreak{}pump}: \code{feed.\allowbreak{}outlet} $\rightarrow$ \code{pump\_\allowbreak{}feed.\allowbreak{}inlet}; \code{feed\_\allowbreak{}pump\_\allowbreak{}to\_\allowbreak{}mixer}: \code{pump\_\allowbreak{}feed.\allowbreak{}outlet} $\rightarrow$ \code{mixer.\allowbreak{}feed}; \code{mixer\_\allowbreak{}to\_\allowbreak{}heat\_\allowbreak{}exchanger\_\allowbreak{}cold\_\allowbreak{}side}: \code{mixer.\allowbreak{}outlet} $\rightarrow$ \code{hx.\allowbreak{}cold\_\allowbreak{}inlet}; \code{heat\_\allowbreak{}exchanger\_\allowbreak{}cold\_\allowbreak{}out\_\allowbreak{}to\_\allowbreak{}brine\_\allowbreak{}pump}: \code{hx.\allowbreak{}cold\_\allowbreak{}outlet} $\rightarrow$ \code{pump\_\allowbreak{}brine.\allowbreak{}inlet}; \code{brine\_\allowbreak{}pump\_\allowbreak{}to\_\allowbreak{}heater}: \code{pump\_\allowbreak{}brine.\allowbreak{}outlet} $\rightarrow$ \code{heater.\allowbreak{}inlet}; \code{heater\_\allowbreak{}to\_\allowbreak{}md\_\allowbreak{}hot\_\allowbreak{}inlet}: \code{heater.\allowbreak{}outlet} $\rightarrow$ \code{MD.\allowbreak{}hot\_\allowbreak{}ch\_\allowbreak{}inlet}; \code{md\_\allowbreak{}hot\_\allowbreak{}outlet\_\allowbreak{}to\_\allowbreak{}concentrate\_\allowbreak{}separator}: \code{MD.\allowbreak{}hot\_\allowbreak{}ch\_\allowbreak{}outlet} $\rightarrow$ \code{separator\_\allowbreak{}concentrate.\allowbreak{}inlet}; \code{concentrate\_\allowbreak{}separator\_\allowbreak{}to\_\allowbreak{}reject}: \code{separator\_\allowbreak{}concentrate.\allowbreak{}reject} $\rightarrow$ \code{reject.\allowbreak{}inlet}; \code{concentrate\_\allowbreak{}separator\_\allowbreak{}recycle\_\allowbreak{}to\_\allowbreak{}mixer}: \code{separator\_\allowbreak{}concentrate.\allowbreak{}recycle} $\rightarrow$ \code{mixer.\allowbreak{}recycle}; \code{chiller\_\allowbreak{}to\_\allowbreak{}md\_\allowbreak{}cold\_\allowbreak{}inlet}: \code{chiller.\allowbreak{}outlet} $\rightarrow$ \code{MD.\allowbreak{}cold\_\allowbreak{}ch\_\allowbreak{}inlet}; \code{md\_\allowbreak{}cold\_\allowbreak{}outlet\_\allowbreak{}to\_\allowbreak{}heat\_\allowbreak{}exchanger\_\allowbreak{}hot\_\allowbreak{}side}: \code{MD.\allowbreak{}cold\_\allowbreak{}ch\_\allowbreak{}outlet} $\rightarrow$ \code{hx.\allowbreak{}hot\_\allowbreak{}inlet}; \code{heat\_\allowbreak{}exchanger\_\allowbreak{}hot\_\allowbreak{}out\_\allowbreak{}to\_\allowbreak{}permeate\_\allowbreak{}separator}: \code{hx.\allowbreak{}hot\_\allowbreak{}outlet} $\rightarrow$ \code{separator\_\allowbreak{}permeate.\allowbreak{}inlet}; \code{permeate\_\allowbreak{}separator\_\allowbreak{}to\_\allowbreak{}product}: \code{separator\_\allowbreak{}permeate.\allowbreak{}permeate} $\rightarrow$ \code{permeate.\allowbreak{}inlet}; \code{permeate\_\allowbreak{}separator\_\allowbreak{}cold\_\allowbreak{}loop\_\allowbreak{}to\_\allowbreak{}pump}: \code{separator\_\allowbreak{}permeate.\allowbreak{}cold\_\allowbreak{}loop\_\allowbreak{}stream} $\rightarrow$ \code{pump\_\allowbreak{}permeate.\allowbreak{}inlet}; \code{permeate\_\allowbreak{}pump\_\allowbreak{}to\_\allowbreak{}chiller}: \code{pump\_\allowbreak{}permeate.\allowbreak{}outlet} $\rightarrow$ \code{chiller.\allowbreak{}inlet}. Terminal feeds: none;
terminal products: none.}

\tracestage{5}{Specification: close operating choices and degrees of freedom}

\tracetool{Topology Validator / Prebuilder}{Specification Agent}{The typed unit,
package, configuration, port, arc, terminal, and construction checks pass for the accepted
TopologyIR; only the accepted topology advances.}

\traceagent{Specification Agent}{DOF Checker}{Emit all 10 persisted SpecIR
assignments one-to-one. The provenance marker \emph{ChE-agent/DoF-validated} denotes a
Specification Agent proposal accepted by the deterministic DoF gate.
\textbf{Unit configuration (8)}:
\code{system.\allowbreak{}overall\_\allowbreak{}recovery} = \code{0.\allowbreak{}5} dimensionless (official default overall water recovery; ChE-agent/DoF-validated); \code{MD.\allowbreak{}permeability\allowbreak{}\_\allowbreak{}coef} = \code{1e-\allowbreak{}10} kg/m/s/Pa (official membrane permeability coefficient; ChE-agent/DoF-validated); \code{MD.\allowbreak{}membrane\_\allowbreak{}thickness} = \code{0.\allowbreak{}0001} m (official membrane thickness; ChE-agent/DoF-validated); \code{MD.\allowbreak{}membrane\_\allowbreak{}thermal\_\allowbreak{}conductivity\allowbreak{}} = \code{0.\allowbreak{}2} W/m/K (official membrane thermal conductivity; ChE-agent/DoF-validated); \code{hx.\allowbreak{}overall\_\allowbreak{}heat\_\allowbreak{}transfer\_\allowbreak{}coefficient} = \code{2000} W/m\textasciicircum{}2/K (official heat exchanger coefficient; ChE-agent/DoF-validated); \code{heater.\allowbreak{}max\_\allowbreak{}outlet\_\allowbreak{}temperature} = \code{363.\allowbreak{}15} K (optimization upper bound from official case; ChE-agent/DoF-validated); \code{chiller.\allowbreak{}min\_\allowbreak{}outlet\_\allowbreak{}temperature} = \code{283.\allowbreak{}15} K (optimization lower bound from official case; ChE-agent/DoF-validated); \code{optimization\allowbreak{}.\allowbreak{}objective} = \code{LCOW} \$/m\textasciicircum{}3 (final WaterTAP optimization objective; ChE-agent/DoF-validated). \textbf{Stream configuration (2)}:
\code{feed.\allowbreak{}feed\_\allowbreak{}flow\_\allowbreak{}mass} = \code{1} kg/s (official default total feed mass flow; ChE-agent/DoF-validated); \code{feed.\allowbreak{}feed\_\allowbreak{}mass\_\allowbreak{}frac\_\allowbreak{}TDS} = \code{0.\allowbreak{}035} mass fraction (official default feed salinity; ChE-agent/DoF-validated). No assignment is replaced by a count.}

\tracetool{DOF Checker}{BuildPlan Compiler}{Validate target existence, compatibility,
duplicate/conflicting fixes, terminal closure, and expected steady-state DoF. The promoted
SpecIR and deterministic DoF report pass before compilation.}

\tracestage{6}{Build and initialize the accepted white-box model}

\tracetool{BuildPlan Compiler}{Initialization / Native Solver}{Compile the hermetic
BuildPlan from the accepted TopologyIR and SpecIR. Initialization contract:
solve order=[feed, MD, concentrate separator, feed pump, mixer, heat exchanger, brine pump, heater, permeate separator, permeate pump, chiller, LCOW optimization]; strategy=WaterTAP MD build, set\_operating\_conditions, initialize\_system through hot and cold recycle loops, simulation solve, then optimize\_set\_up and LCOW solve. Recycle/tear contract: []. Solve contract:
native\_solve\_disabled=false; expected\_dof=0 for simulation, 6 for optimization.}

\tracestage{7}{Solve, inspect diagnostics, repair, and optimize only when eligible}

\tracetool{Native Solver}{Failure Router / Report}{Persist pass=\code{True},
termination \code{optimal}, final DoF
\code{0}, stage \code{steady\_\allowbreak{}state\_\allowbreak{}simulation}, and no terminal error.
Representative same-run stream/product quantities: \code{heat\_\allowbreak{}exchanger\_\allowbreak{}hot\_\allowbreak{}out\_\allowbreak{}to\_\allowbreak{}permeate\_\allowbreak{}separator} [mass\_flow=8.04107; flow\_kg\_s=8.04107; temperature=298.03; pressure=374,271; flow\_mass\_phase\_comp: Liq/H2O=8.04107, Liq/TDS=4.47114e-17]; \code{permeate\_\allowbreak{}separator\_\allowbreak{}to\_\allowbreak{}product} [mass\_flow=0.4825; flow\_kg\_s=0.4825; temperature=298.03; pressure=374,271; flow\_mass\_phase\_comp: Liq/H2O=0.4825, Liq/TDS=5.91586e-20]; \code{permeate\_\allowbreak{}separator\_\allowbreak{}cold\_\allowbreak{}loop\_\allowbreak{}to\_\allowbreak{}pump} [mass\_flow=7.55857; flow\_kg\_s=7.55857; temperature=298.03; pressure=374,271; flow\_mass\_phase\_comp: Liq/H2O=7.55857, Liq/TDS=4.47114e-17]; \code{permeate\_\allowbreak{}pump\_\allowbreak{}to\_\allowbreak{}chiller} [mass\_flow=7.55857; flow\_kg\_s=7.55857; temperature=298.03; pressure=601,325; flow\_mass\_phase\_comp: Liq/H2O=7.55857, Liq/TDS=4.47114e-17]. Source
evidence is reported as the accepted output of the native execution node.}

\traceagent{Debug Supervisor audit (standby)}{Failure Router / Report}{No fault was routed to
Debug; no diagnosis, repair artifact, or replay was generated. The supervisor remained armed
from before Stage 1 through terminal acceptance.}

\tracetool{Failure Router}{Optimization Agent / Report}{Read the structured topology,
specification, DoF, and accepted execution reports. Close the bounded repair branch, then pass the
accepted simulator state to Optimization only when the request makes that role eligible.}

\traceagent{Optimization Agent}{Deterministic runner / Report}{Optimization was invoked through the collaborating agent plan and deterministic runner. Objective: minimize levelized cost of water. Validated variables: \code{fs.\allowbreak{}MD.\allowbreak{}area}; \code{fs.\allowbreak{}MD.\allowbreak{}length}; \code{fs.\allowbreak{}hx.\allowbreak{}area}; \code{fs.\allowbreak{}heater.\allowbreak{}control\_\allowbreak{}volume.\allowbreak{}properties\_\allowbreak{}out[\allowbreak{}0]\allowbreak{}.\allowbreak{}temperature}; \code{fs.\allowbreak{}chiller.\allowbreak{}control\_\allowbreak{}volume.\allowbreak{}properties\_\allowbreak{}out[\allowbreak{}0]\allowbreak{}.\allowbreak{}temperature}; \code{fs.\allowbreak{}pump\_\allowbreak{}brine.\allowbreak{}control\_\allowbreak{}volume.\allowbreak{}properties\_\allowbreak{}out[\allowbreak{}0]\allowbreak{}.\allowbreak{}pressure}; target DoF \code{6}; plan constraints: MD hot-channel inlet pressure <= 1e6 Pa. Native same-run outcome: pass=true; stage=design\_optimization\_solver; case\_family=watertap\_membrane\_distillation; official\_reference\_url=https://watertap.readthedocs.io/en/stable/technical\_reference/flowsheets/membrane\_distillation.html; simulation\_termination\_condition=optimal; termination\_condition=optimal; final\_dof=6; water\_recovery=0.5; recycle\_ratio=6.1696; levelized\_cost\_of\_water=15.4446; specific\_energy\_consumption=182.872; thermal\_efficiency=0.678341; effectiveness=0.858913; permeate\_tds\_mass\_fraction=-4.0516e-38; md\_area\_m2=103.182; heat\_exchanger\_area\_m2=291.078; heater\_outlet\_temperature\_k=363.15; chiller\_outlet\_temperature\_k=291.629; brine\_pump\_outlet\_pressure\_pa=1e+06. Decision-variable values: . Constraint values: . Product/design values: .}

\tracetool{IDAES FV exporter}{Web / manuscript report}{After solve and any eligible
optimization, export the same-run flowsheet visualization with 28 FV
cells. Retain the ChE process evidence, accepted topology, FV graph, and terminal numerical
evidence as one LangGraph trajectory.}

\end{tracecasebox}

\casefigure{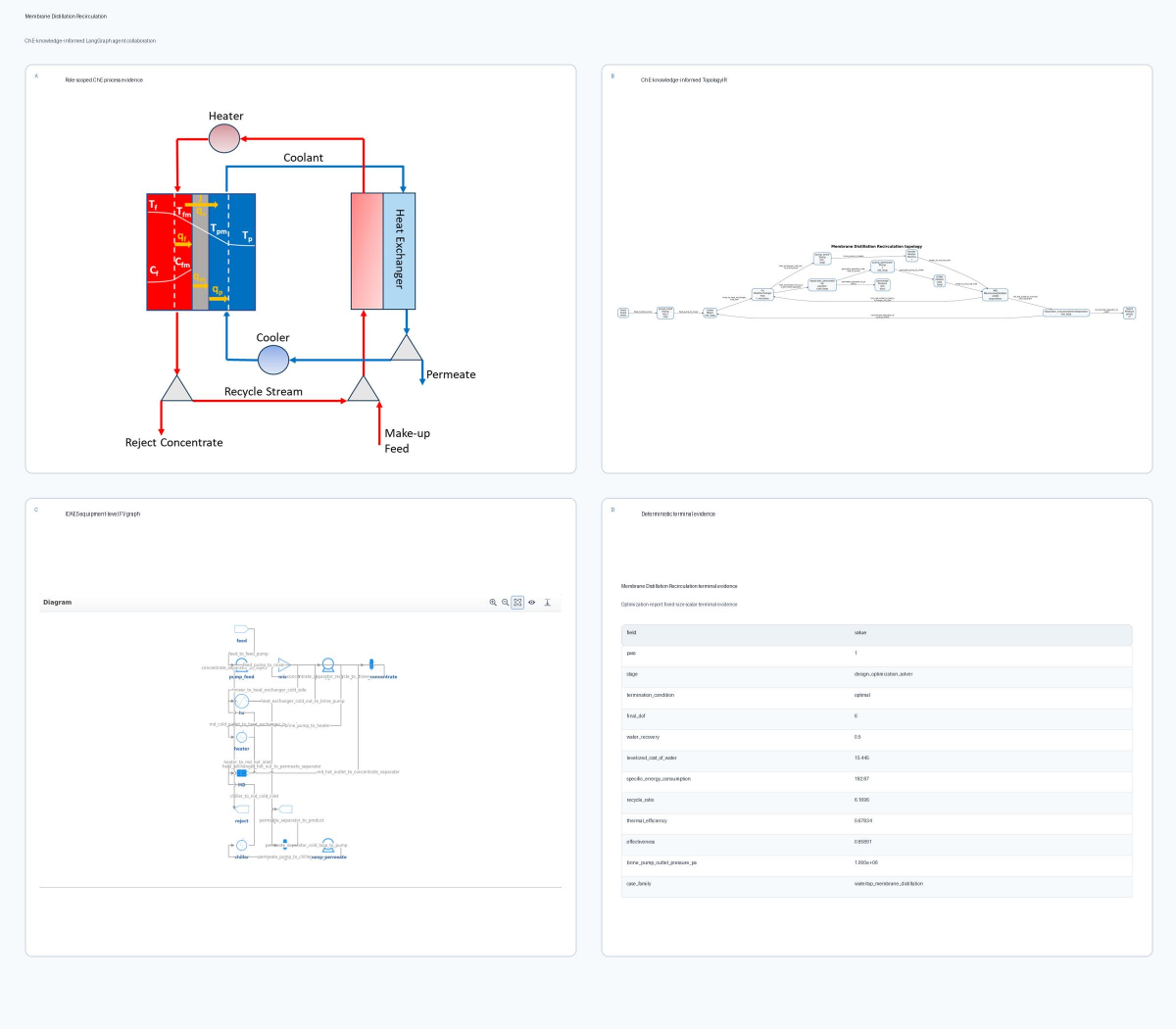}
{WaterTAP membrane-distillation recirculation collaboration audit bundle.}
{fig:arxiv-case-md}

\subsubsection{Autothermal Reformer}
\begin{tracecasebox}{Autothermal reformer---complete interaction trace}
\scriptsize
\runbar{}

\tracetool{LangGraph collaboration profile}{Case report}{Input Understanding, Intent,
Visual, Topology, Specification, deterministic validation, execution, and conditional
Debug/Optimization roles form one artifact-consistent trajectory. Role reasoning is bounded by
medium. The visibility gateway records 6 scoped model-agent
calls and 18 visibility-audit events with zero boundary errors.
Structured-output repairs=0; Debug replays=0. The
Debug supervisor is armed before Stage 1. The deterministic execution returns \code{optimized} with
final acceptance at the report boundary.}

\tracestage{1}{Input Understanding: preserve requirements, relations, and negations}

\traceuser{Root User}{Input Understanding}{Build the NGFC autothermal-reformer subsystem with staged air compression and intercooling, natural-gas expansion, recuperative heating, feed and bypass mixing, autothermal reforming, and the downstream bypass rejoin. Include the steady-state design optimization.}

\traceagent{Input Understanding}{Intent Router}{Normalize the request as
\code{optimize} in Natural Gas Fuel Cell (NGFC) preprocessing. Required
features: \code{staged\_\allowbreak{}air\_\allowbreak{}compression}; \code{intercooling\allowbreak{}}; \code{natural\_\allowbreak{}gas\_\allowbreak{}expansion}; \code{recuperative\allowbreak{}\_\allowbreak{}heater}; \code{feed\_\allowbreak{}bypass\_\allowbreak{}mixer}; \code{autothermal\_\allowbreak{}reformer}; \code{bypass\_\allowbreak{}rejoin\_\allowbreak{}mixer}. Negated features:
none. Directed relations: air\_compressor\_stage\_1 feeds intercooler; intercooler feeds air\_compressor\_stage\_2; natural\_gas\_inlet feeds expansion\_valve; expansion\_valve feeds recuperative\_heater\_cold\_side; recuperative\_heater\_hot\_side feeds autothermal\_reformer; feed\_bypass\_mixer feeds autothermal\_reformer; autothermal\_reformer feeds bypass\_rejoin\_mixer.
Assumptions: Assume standard NGFC preprocessing topology where air and fuel streams are prepared separately before mixing or reacting.;
ambiguities: none;
clarification required=\code{False}; confidence
0.95.}

\tracestage{2}{Intent: recover the ChE process boundary and downstream obligations}

\traceagent{Intent Router}{ChE evidence coordinator / Visual Descriptor}{Interpret the
request as NGFC ATR in the process\_system domain.
Route the required equipment and process features---\code{staged\_\allowbreak{}air\_\allowbreak{}compression}; \code{intercooling\allowbreak{}}; \code{natural\_\allowbreak{}gas\_\allowbreak{}expansion}; \code{recuperative\allowbreak{}\_\allowbreak{}heater}; \code{feed\_\allowbreak{}bypass\_\allowbreak{}mixer}; \code{autothermal\_\allowbreak{}reformer}; \code{bypass\_\allowbreak{}rejoin\_\allowbreak{}mixer}---
and directed ChE relations---air\_compressor\_stage\_1 feeds intercooler; intercooler feeds air\_compressor\_stage\_2; natural\_gas\_inlet feeds expansion\_valve; expansion\_valve feeds recuperative\_heater\_cold\_side; recuperative\_heater\_hot\_side feeds autothermal\_reformer; feed\_bypass\_mixer feeds autothermal\_reformer; autothermal\_reformer feeds bypass\_rejoin\_mixer---to the downstream roles. Preserve explicit
negations and bounded assumptions; confidence 1.}

\tracestage{3}{Visual evidence: translate the PFD into an inspectable observation}

\tracetool{ChE evidence coordinator}{Visual Descriptor}{Provide a role-scoped engineering
evidence packet from the process request and authoritative process documentation. It contains
equipment functions, visible labels, material/energy-flow direction, recycle or branch cues,
operating semantics, and an explicit uncertainty boundary. LangGraph exposes this ChE evidence
to the perception role without adding executable implementation details.}

\traceagent{Visual Descriptor}{VisualGraphIR Validator / Topology Agent}{
Act as the modeling-assistance perception adapter. The persisted
VisualGraphIR contains 38 nodes and
14 edges. Its actual observation excerpt is retained below:
\begin{tracejsonbox}
\{\\
  "schema\_version": "visual\_graph\_ir/2",\\
  "nodes": [\\
    \{\\
      "visual\_id": "Fs\_air\_compressor\_s1\_inlet",\\
      "label\_text": "Fs.air\_compressor\_s1.inlet",\\
      "equipment\_class\_guess": "PressureChanger",\\
      "bbox": [\\
        0.18,\\
        0.27,\\
        0.32,\\
        0.35\\
      ],\\
      "confidence\_0\_1": 0.95,\\
      "evidence\_ref": "Top-left labeled inlet arrow for air compressor stage 1"\\
    \}\\
  ],\\
  "edges": [\\
    \{\\
      "visual\_edge\_id": "TO\_NG\_EXP",\\
      "source\_visual\_id": "recuperator\_shell\_outlet",\\
      "destination\_visual\_id": "natural\_gas\_expander\_inlet",\\
      "label\_text": "Preheated Natural Gas",\\
      "direction\_confidence\_0\_1": 0.95,\\
      "is\_recycle\_guess": false,\\
      "evidence\_ref": "Dashed purple arrow from recuperator to expander"\\
    \}\\
  ]\\
\}\\
\end{tracejsonbox}
A one-to-one audit against the same run's source flowchart found a bounded visible-graph
correction: Collapse endpoint-only port placeholders back onto the visible equipment to restore a connected equipment graph; Retain the air-compression train, steam and natural-gas feeds, reformer bypass, recuperator, and final syngas/bypass rejoin shown in the source. This is an
explicit source-verification supplement and is not reattributed to the original Visual
Descriptor response:
\begin{tracejsonbox}
\{\\
  "audit\_status": "same-source publication correction",\\
  "mode": "replace",\\
  "verified\_node\_count": 14,\\
  "verified\_edge\_count": 15,\\
  "correction\_summary": [\\
    "Collapse endpoint-only port placeholders back onto the visible equipment to restore a connected equipment graph.",\\
    "Retain the air-compression train, steam and natural-gas feeds, reformer bypass, recuperator, and final syngas/bypass rejoin shown in the source."\\
  ]\\
\}\\
\end{tracejsonbox}
The source-verified publication layer contains 14 nodes and
15 edges. Complete node inventory:
\code{AirFeed}=Feed; \code{air\_\allowbreak{}compressor\_\allowbreak{}s1}=PressureChanger; \code{intercooler\_\allowbreak{}s1}=Heater; \code{air\_\allowbreak{}compressor\_\allowbreak{}s2}=PressureChanger; \code{intercooler\_\allowbreak{}s2}=Heater; \code{SteamFeed}=Feed; \code{NaturalGasFe\allowbreak{}ed}=Feed; \code{reformer\_\allowbreak{}recuperator}=HeatExchanger; \code{natural\_\allowbreak{}gas\_\allowbreak{}expander}=PressureChanger; \code{reformer\_\allowbreak{}bypass}=Separator; \code{reformer\_\allowbreak{}mix}=Mixer; \code{reformer}=GibbsReactor; \code{bypass\_\allowbreak{}rejoin}=Mixer; \code{ToPowerIslan\allowbreak{}d}=Product. Complete directed-edge inventory: \code{air\_\allowbreak{}feed\_\allowbreak{}to\_\allowbreak{}c1}: \code{AirFeed}$\rightarrow$\code{air\_\allowbreak{}compressor\_\allowbreak{}s1}; \code{c1\_\allowbreak{}to\_\allowbreak{}ic1}: \code{air\_\allowbreak{}compressor\_\allowbreak{}s1}$\rightarrow$\code{intercooler\_\allowbreak{}s1}; \code{ic1\_\allowbreak{}to\_\allowbreak{}c2}: \code{intercooler\_\allowbreak{}s1}$\rightarrow$\code{air\_\allowbreak{}compressor\_\allowbreak{}s2}; \code{c2\_\allowbreak{}to\_\allowbreak{}ic2}: \code{air\_\allowbreak{}compressor\_\allowbreak{}s2}$\rightarrow$\code{intercooler\_\allowbreak{}s2}; \code{ic2\_\allowbreak{}to\_\allowbreak{}mix}: \code{intercooler\_\allowbreak{}s2}$\rightarrow$\code{reformer\_\allowbreak{}mix}; \code{steam\_\allowbreak{}to\_\allowbreak{}mix}: \code{SteamFeed}$\rightarrow$\code{reformer\_\allowbreak{}mix}; \code{ng\_\allowbreak{}to\_\allowbreak{}recuperator}: \code{NaturalGasFe\allowbreak{}ed}$\rightarrow$\code{reformer\_\allowbreak{}recuperator}; \code{recuperator\_\allowbreak{}to\_\allowbreak{}expander}: \code{reformer\_\allowbreak{}recuperator}$\rightarrow$\code{natural\_\allowbreak{}gas\_\allowbreak{}expander}; \code{expander\_\allowbreak{}to\_\allowbreak{}bypass}: \code{natural\_\allowbreak{}gas\_\allowbreak{}expander}$\rightarrow$\code{reformer\_\allowbreak{}bypass}; \code{bypass\_\allowbreak{}to\_\allowbreak{}mix}: \code{reformer\_\allowbreak{}bypass}$\rightarrow$\code{reformer\_\allowbreak{}mix}; \code{mix\_\allowbreak{}to\_\allowbreak{}reformer}: \code{reformer\_\allowbreak{}mix}$\rightarrow$\code{reformer}; \code{reformer\_\allowbreak{}to\_\allowbreak{}recuperator}: \code{reformer}$\rightarrow$\code{reformer\_\allowbreak{}recuperator}; \code{recuperator\_\allowbreak{}to\_\allowbreak{}rejoin}: \code{reformer\_\allowbreak{}recuperator}$\rightarrow$\code{bypass\_\allowbreak{}rejoin}; \code{bypass\_\allowbreak{}to\_\allowbreak{}rejoin}: \code{reformer\_\allowbreak{}bypass}$\rightarrow$\code{bypass\_\allowbreak{}rejoin}; \code{rejoin\_\allowbreak{}to\_\allowbreak{}product}: \code{bypass\_\allowbreak{}rejoin}$\rightarrow$\code{ToPowerIslan\allowbreak{}d}. Visible text labels:
none.
Uncertainty boundary: The validator represented 23 one-sided visual edge endpoints as endpoint-only boundary nodes (11 feed-side and 12 product-side). This representation-only closure added no edge or internal equipment. The Topology Agent consumes the persisted promoted artifact together
with routed ChE constraints and simulator capabilities; the audit layer documents the
flowchart-to-VisualIR correction transparently rather than altering the selected run.}

\tracetool{VisualGraphIR Validator}{Topology Agent}{Validate the visual schema, unique
node/edge identifiers, normalized geometry, endpoint closure, evidence references, and
uncertainty boundary before topology synthesis.}

\tracestage{4}{Topology: combine observations with ChE and simulator constraints}

\traceagent{Topology Agent}{Topology Validator / Prebuilder}{Combine the typed visual record
with routed ChE constraints and simulator capabilities. Emit all
10 typed units: \code{reformer\_\allowbreak{}recuperator} HeatExchanger (ChE role preheat natural gas using hot reformer outlet, package \code{NG\_\allowbreak{}props}, stage \code{atr\_\allowbreak{}reformer}); \code{NG\_\allowbreak{}expander} PressureChanger (ChE role expand preheated natural gas before bypass split, package \code{NG\_\allowbreak{}props}, stage \code{atr\_\allowbreak{}reformer}); \code{reformer\_\allowbreak{}bypass} Separator (ChE role split natural gas between reformer and bypass, package \code{NG\_\allowbreak{}props}, stage \code{atr\_\allowbreak{}reformer}); \code{air\_\allowbreak{}compressor\_\allowbreak{}s1} PressureChanger (ChE role first-stage air compressor, package \code{NG\_\allowbreak{}props}, stage \code{atr\_\allowbreak{}reformer}); \code{intercooler\_\allowbreak{}s1} Heater (ChE role first intercooler, package \code{NG\_\allowbreak{}props}, stage \code{atr\_\allowbreak{}reformer}); \code{air\_\allowbreak{}compressor\_\allowbreak{}s2} PressureChanger (ChE role second-stage air compressor, package \code{NG\_\allowbreak{}props}, stage \code{atr\_\allowbreak{}reformer}); \code{intercooler\_\allowbreak{}s2} Heater (ChE role second intercooler, package \code{NG\_\allowbreak{}props}, stage \code{atr\_\allowbreak{}reformer}); \code{reformer\_\allowbreak{}mix} Mixer (ChE role mix natural gas, compressed air, and steam, package \code{NG\_\allowbreak{}props}, stage \code{atr\_\allowbreak{}reformer}); \code{reformer} GibbsReactor (ChE role autothermal reforming Gibbs reactor, package \code{NG\_\allowbreak{}props}, stage \code{atr\_\allowbreak{}reformer}); \code{bypass\_\allowbreak{}rejoin} Mixer (ChE role mix reformer syngas with natural-gas bypass, package \code{NG\_\allowbreak{}props}, stage \code{atr\_\allowbreak{}reformer}). Property/reaction package
surface: \code{NG\_\allowbreak{}props} (components=[H2, CO, H2O, CO2, CH4, C2H6, C3H8, C4H10, N2, O2, Ar]; phases=[Vap]; scope=[reformer\_recuperator, NG\_expander, reformer\_bypass, air\_compressor\_s1, intercooler\_s1, air\_compressor\_s2, intercooler\_s2, reformer\_mix, reformer, bypass\_rejoin]). Emit all 11 exact stream endpoints:
\code{TO\_\allowbreak{}NG\_\allowbreak{}EXP}: \code{reformer\_\allowbreak{}recuperator.\allowbreak{}tube\_\allowbreak{}outlet} $\rightarrow$ \code{NG\_\allowbreak{}expander.\allowbreak{}inlet}; \code{NG\_\allowbreak{}EXP\_\allowbreak{}OUT}: \code{NG\_\allowbreak{}expander.\allowbreak{}outlet} $\rightarrow$ \code{reformer\_\allowbreak{}bypass.\allowbreak{}inlet}; \code{TO\_\allowbreak{}REF}: \code{reformer\_\allowbreak{}bypass.\allowbreak{}reformer\_\allowbreak{}outlet} $\rightarrow$ \code{reformer\_\allowbreak{}mix.\allowbreak{}gas\_\allowbreak{}inlet}; \code{STAGE\_\allowbreak{}1\_\allowbreak{}OUT}: \code{air\_\allowbreak{}compressor\_\allowbreak{}s1.\allowbreak{}outlet} $\rightarrow$ \code{intercooler\_\allowbreak{}s1.\allowbreak{}inlet}; \code{IC\_\allowbreak{}1\_\allowbreak{}OUT}: \code{intercooler\_\allowbreak{}s1.\allowbreak{}outlet} $\rightarrow$ \code{air\_\allowbreak{}compressor\_\allowbreak{}s2.\allowbreak{}inlet}; \code{STAGE\_\allowbreak{}2\_\allowbreak{}OUT}: \code{air\_\allowbreak{}compressor\_\allowbreak{}s2.\allowbreak{}outlet} $\rightarrow$ \code{intercooler\_\allowbreak{}s2.\allowbreak{}inlet}; \code{IC\_\allowbreak{}2\_\allowbreak{}OUT}: \code{intercooler\_\allowbreak{}s2.\allowbreak{}outlet} $\rightarrow$ \code{reformer\_\allowbreak{}mix.\allowbreak{}oxygen\_\allowbreak{}inlet}; \code{REF\_\allowbreak{}IN}: \code{reformer\_\allowbreak{}mix.\allowbreak{}outlet} $\rightarrow$ \code{reformer.\allowbreak{}inlet}; \code{REF\_\allowbreak{}OUT}: \code{reformer.\allowbreak{}outlet} $\rightarrow$ \code{reformer\_\allowbreak{}recuperator.\allowbreak{}shell\_\allowbreak{}inlet}; \code{REF\_\allowbreak{}RECUP\_\allowbreak{}OUT}: \code{reformer\_\allowbreak{}recuperator.\allowbreak{}shell\_\allowbreak{}outlet} $\rightarrow$ \code{bypass\_\allowbreak{}rejoin.\allowbreak{}syngas\_\allowbreak{}inlet}; \code{REF\_\allowbreak{}BYPASS}: \code{reformer\_\allowbreak{}bypass.\allowbreak{}bypass\_\allowbreak{}outlet} $\rightarrow$ \code{bypass\_\allowbreak{}rejoin.\allowbreak{}bypass\_\allowbreak{}inlet}. Terminal feeds: \code{reformer\_\allowbreak{}recuperator.\allowbreak{}tube\_\allowbreak{}inlet}; \code{air\_\allowbreak{}compressor\_\allowbreak{}s1.\allowbreak{}inlet}; \code{reformer\_\allowbreak{}mix.\allowbreak{}steam\_\allowbreak{}inlet};
terminal products: \code{bypass\_\allowbreak{}rejoin.\allowbreak{}outlet}.}

\tracestage{5}{Specification: close operating choices and degrees of freedom}

\tracetool{Topology Validator / Prebuilder}{Specification Agent}{The typed unit,
package, configuration, port, arc, terminal, and construction checks pass for the accepted
TopologyIR; only the accepted topology advances.}

\traceagent{Specification Agent}{DOF Checker}{Emit all 7 persisted SpecIR
assignments one-to-one. The provenance marker \emph{ChE-agent/DoF-validated} denotes a
Specification Agent proposal accepted by the deterministic DoF gate.
\textbf{Unit configuration (2)}:
\code{reformer\_\allowbreak{}bypass.\allowbreak{}split\_\allowbreak{}fraction[\allowbreak{}0, 'bypass\_\allowbreak{}outlet']\allowbreak{}} = \code{0.\allowbreak{}6} dimensionless (internal reformation bypass fraction; ChE-agent/DoF-validated); \code{reformer\_\allowbreak{}mix.\allowbreak{}steam\_\allowbreak{}to\_\allowbreak{}ng\_\allowbreak{}ratio} = \code{1} dimensionless (official steam flow constraint uses steam = (1 - IR) * NG flow; ChE-agent/DoF-validated). \textbf{Stream configuration (5)}:
\code{reformer\_\allowbreak{}recuperator.\allowbreak{}tube\_\allowbreak{}inlet.\allowbreak{}flow\_\allowbreak{}mol[\allowbreak{}0]\allowbreak{}} = \code{1161.\allowbreak{}9} mol/s (natural gas feed flow; ChE-agent/DoF-validated); \code{air\_\allowbreak{}compressor\_\allowbreak{}s1.\allowbreak{}inlet.\allowbreak{}flow\_\allowbreak{}mol[\allowbreak{}0]\allowbreak{}} = \code{1332.\allowbreak{}9} mol/s (air feed initial flow; ChE-agent/DoF-validated); \code{reformer\_\allowbreak{}mix.\allowbreak{}steam\_\allowbreak{}inlet.\allowbreak{}flow\_\allowbreak{}mol[\allowbreak{}0]\allowbreak{}} = \code{464.\allowbreak{}77} mol/s (steam feed initial flow; ChE-agent/DoF-validated); \code{reformer.\allowbreak{}outlet.\allowbreak{}temperature[\allowbreak{}0]\allowbreak{}} = \code{1060.\allowbreak{}93} K (reformer outlet temperature; ChE-agent/DoF-validated); \code{reformer.\allowbreak{}outlet.\allowbreak{}pressure[\allowbreak{}0]\allowbreak{}} = \code{137895} Pa (reformer outlet pressure; ChE-agent/DoF-validated). No assignment is replaced by a count.}

\tracetool{DOF Checker}{BuildPlan Compiler}{Validate target existence, compatibility,
duplicate/conflicting fixes, terminal closure, and expected steady-state DoF. The promoted
SpecIR and deterministic DoF report pass before compilation.}

\tracestage{6}{Build and initialize the accepted white-box model}

\tracetool{BuildPlan Compiler}{Initialization / Native Solver}{Compile the hermetic
BuildPlan from the accepted TopologyIR and SpecIR. Initialization contract:
solve order=[reformer, reformer\_recuperator, NG\_expander, reformer\_bypass, air\_compressor\_s1, intercooler\_s1, air\_compressor\_s2, intercooler\_s2, reformer\_mix, bypass\_rejoin]; strategy=simulator-native NGFC initialize\_reformer(). Recycle/tear contract: []. Solve contract:
native\_solve\_disabled=false; expected\_dof=0.}

\tracestage{7}{Solve, inspect diagnostics, repair, and optimize only when eligible}

\tracetool{Native Solver}{Failure Router / Report}{Persist pass=\code{True},
termination \code{optimal}, final DoF
\code{0}, stage \code{steady\_\allowbreak{}state\_\allowbreak{}solver}, and no terminal error.
Representative same-run stream/product quantities: \code{steam\_\allowbreak{}feed} [flow\_mol=464.76; temperature=422; pressure=206,843; mole\_frac\_comp: H2=1e-11, CO=1e-11, H2O=1, CO2=1e-11]; \code{reformer\_\allowbreak{}inlet} [flow\_mol=2262.45; temperature=469.838; pressure=203,396; mole\_frac\_comp: H2=2.59655e-11, CO=2.59655e-11, H2O=0.21155, CO2=0.00223098]; \code{reformer\_\allowbreak{}outlet} [flow\_mol=2942.07; temperature=1060.93; pressure=137,895; mole\_frac\_comp: H2=0.34036, CO=0.112478, H2O=0.13826, CO2=0.0517876]; \code{atr\_\allowbreak{}syngas\_\allowbreak{}product} [flow\_mol=3639.21; temperature=621.322; pressure=137,895; mole\_frac\_comp: H2=0.27516, CO=0.0909317, H2O=0.111775, CO2=0.0437826]. Source
evidence is reported as the accepted output of the native execution node.}

\traceagent{Debug Supervisor audit (standby)}{Failure Router / Report}{No fault was routed to
Debug; no diagnosis, repair artifact, or replay was generated. The supervisor remained armed
from before Stage 1 through terminal acceptance.}

\tracetool{Failure Router}{Optimization Agent / Report}{Read the structured topology,
specification, DoF, and accepted execution reports. Close the bounded repair branch, then pass the
accepted simulator state to Optimization only when the request makes that role eligible.}

\traceagent{Optimization Agent}{Deterministic runner / Report}{Optimization was invoked through the collaborating agent plan and deterministic runner. Objective: maximize syngas H2 mole fraction. Validated variables: \code{fs.\allowbreak{}reformer\_\allowbreak{}bypass.\allowbreak{}split\_\allowbreak{}fraction[\allowbreak{}0, 'bypass\_\allowbreak{}outlet']\allowbreak{}}; target DoF \code{1}; plan constraints: syngas N2 mole fraction <= 0.34. Native same-run outcome: pass=true; stage=steady\_state\_optimization; case\_family=ngfc\_atr; official\_reference\_url=https://idaes.github.io/examples-pse/latest/Examples/Flowsheets/power\_generation/ngfc/NGFC\_flowsheet\_doc.html; optimization\_form=rigorous\_ngfc\_reformer\_subsystem\_ipopt; termination\_condition=optimal; steady\_state\_dof\_before\_optimization=0; optimization\_degrees\_of\_freedom=1; final\_dof=1; objective\_max\_h2\_mole\_fraction=0.327834. Decision-variable values: bypass\_frac=0.194765; steam\_to\_ng\_ratio\_effective=1. Constraint values: bypass\_frac\_bounds=[0.1, 0.8]; max\_product\_n2\_mole\_fraction=0.34; product\_n2\_mole\_fraction=0.34. Product/design values: atr\_syngas\_h2\_mole\_fraction=0.327834; atr\_syngas\_n2\_mole\_fraction=0.34; atr\_syngas\_flow\_mol=6148.94; atr\_syngas\_temperature=815.911; atr\_syngas\_pressure=137,895.}

\tracetool{IDAES FV exporter}{Web / manuscript report}{After solve and any eligible
optimization, export the same-run flowsheet visualization with 29 FV
cells. Retain the ChE process evidence, accepted topology, FV graph, and terminal numerical
evidence as one LangGraph trajectory.}

\end{tracecasebox}

\casefigure{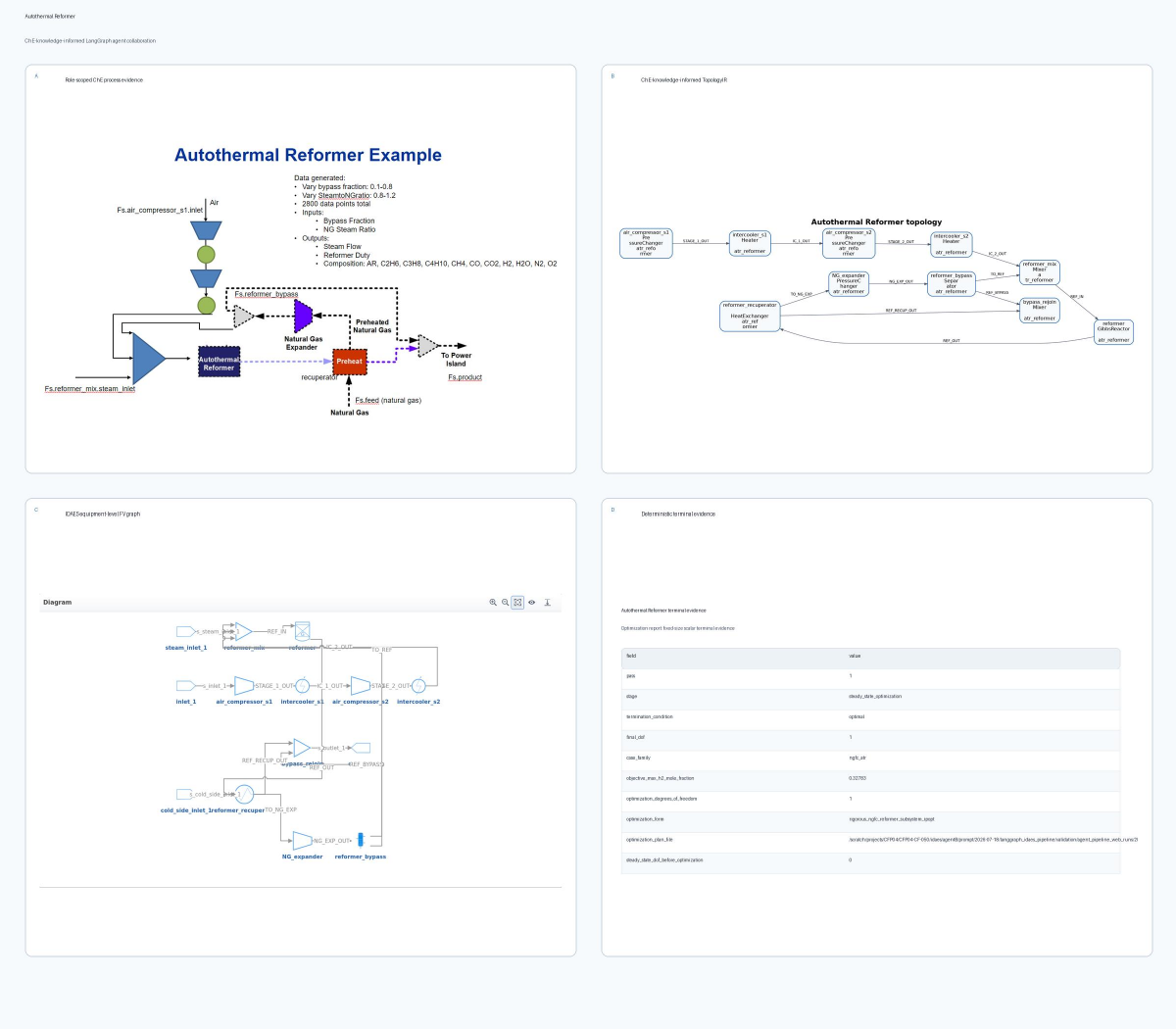}
{Autothermal reformer collaboration audit bundle.}
{fig:arxiv-case-atr}

\clearpage
\bibliographystyle{unsrtnat}
\bibliography{references}

\end{document}